\pdfoutput=1
\documentclass[runningheads]{llncs}
\usepackage[final,year=2026]{eccv}
\usepackage{eccvabbrv}
\usepackage{graphicx}
\usepackage{booktabs}
\usepackage{array}
\usepackage{tabularx}
\usepackage{hyperref}
\hypersetup{hidelinks,pdftitle={Does Marginal Coverage Guarantee Class-Conditional Safety for Zero-Shot VLMs Under Shift?},bookmarksopen=true}
\usepackage{bookmark}
\usepackage{placeins}
\newcounter{suppsec}\renewcommand{\thesuppsec}{S\arabic{suppsec}}
\newcommand{\suppsection}[1]{\refstepcounter{suppsec}%
  \subsection*{\thesuppsec\quad #1}%
  \pdfbookmark[1]{\thesuppsec\ #1}{suppsec.\thesuppsec}}
\newcounter{supppart}
\newcommand{\supppart}[1]{\FloatBarrier\refstepcounter{supppart}\bigskip%
  \begin{center}\large\bfseries Part \Roman{supppart}: #1\end{center}\nopagebreak\medskip%
  \pdfbookmark[0]{Part \Roman{supppart}: #1}{supppart.\thesupppart}}
\begin{document}
\title{Does Marginal Coverage Guarantee Class-Conditional Safety for Zero-Shot VLMs Under Shift?}
\titlerunning{Marginal Coverage Is Not Class-Conditional Safety}
\author{Jai Kumar Sharma\inst{1} \and Amartya Dutta\inst{1}}
\authorrunning{J.~K.~Sharma and A.~Dutta}
\institute{Virginia Tech, Blacksburg, VA, USA\\\email{\{jaisharma,amartya\}@vt.edu}}
\maketitle
\begin{abstract}
Split-conformal prediction provides marginal coverage under exchangeability and is increasingly used as an abstention layer for zero-shot vision-language models (VLMs). We audit this practice under deployment shift for CLIP, OpenCLIP, and SigLIP across ImageNet and non-ImageNet settings. Marginal coverage can remain relatively high while class-conditional tail coverage collapses: on ImageNet-Sketch, worst-class coverage falls to $\approx$0 and 10--12\% of classes lie below a finite-sample null floor, despite marginal coverage of about 0.86. The failure is aligned with target-domain class accuracy but is not predicted by the source-domain diagnostics we test. Source-side Mondrian calibration improves the in-distribution tail but does not transfer, while clustered conformal and Conf-OT improve marginal or average metrics without recovering the worst-class tail. Target-side class calibration substantially lifts the tail, but requires labels for every class and remains set-size-intensive. We further identify a 2--3$\times$ cross-family efficiency gap and show that native SigLIP sigmoid scores remove APS's probability-mass interpretation. The findings persist across the tested model scale, pretraining corpus, prompt, miscoverage level $\alpha$, and shifted non-ImageNet settings. Marginal conformal coverage should therefore be treated as an average reliability statistic, not as a safety guarantee for the class tail.
\end{abstract}
\pdfbookmark[0]{1. Introduction}{mainsec.1}
\section{Introduction}
Split-conformal prediction converts a classifier's scores into prediction sets with a finite-sample, distribution-free marginal coverage guarantee, P(Y $\in$ C(X)) $\geq$ 1-$\alpha$, when calibration and test data are exchangeable [\hyperlink{bib.ref1}{1}, \hyperlink{bib.ref2}{2}]. This assumption-dependent guarantee makes conformal prediction attractive as a deployment-time uncertainty wrapper for zero-shot vision-language models (VLMs): one calibrates a threshold on a labeled source split and returns a set of plausible labels at test time.

In zero-shot VLM deployment, exchangeability is routinely violated. Visual inputs may shift from natural photographs to sketches, renditions, corruptions, or adversarially filtered examples, and the label space can also change, since the candidate classes are defined at inference time by text prompts. Once calibration and test data are no longer exchangeable, the theorem no longer applies; the relevant question is therefore how empirical coverage behaves.

Recent work suggests that adaptive conformal scores can remain effective for CLIP-like models under several ImageNet shifts, at least in terms of marginal coverage [\hyperlink{bib.ref11}{11}, \hyperlink{bib.ref12}{12}]. We show this picture is incomplete. Across CLIP, OpenCLIP, and SigLIP, marginal coverage and set size are family- and score-dependent. More importantly, relatively high marginal coverage can hide severe class-conditional tail failures that are not repaired by the deployable methods we test. These failures are accuracy-aligned, not identified by the source-domain diagnostics we test, and robust across the tested model and dataset axes.

We study this gap as an audit rather than as a new conformal algorithm. The deployment question is whether standard wrappers for frozen, black-box zero-shot VLMs provide the class-tail safety that a practitioner may infer from a marginal coverage number. In our experiments, they do not: some classes receive essentially no coverage even when the marginal number appears acceptable, and the failing classes are not identified by the source-domain diagnostics we test. Throughout, ``safety'' refers to this statistical property---class-conditional coverage at the nominal level---rather than measured downstream task harm.

\textbf{Contributions.} Our contribution is empirical and diagnostic rather than algorithmic: we propose no new conformal method. First, we provide a cross-family audit of conformal prediction for frozen zero-shot VLMs, comparing CLIP, OpenCLIP, and SigLIP under visual, corruption, and vocabulary shift. The audit reveals family-dependent marginal coverage, a 2--3$\times$ set-size gap, and a score-geometry mismatch between APS and native SigLIP sigmoid scores. Second, we show that marginal coverage hides class-conditional tail collapse: a finite-sample null analysis demonstrates that the failure extends beyond worst-of-$K$ sampling noise, and source-side Mondrian thresholds do not transfer to the target. Third, we evaluate the available repair and deployment options. Source-Mondrian, clustered conformal, and Conf-OT improve marginal or average metrics but not the worst-class tail; target-side class calibration lifts the tail at high label and set-size cost, while label-free selective prediction abstains from rather than covers the hard tail. We analyze the underlying mechanism and test the conclusions across scale, corpus, prompt, $\alpha$, and non-ImageNet settings.

\pdfbookmark[0]{2. Related work}{mainsec.2}
\section{Related work}
\textbf{Conformal prediction.} Split-conformal prediction gives finite-sample marginal coverage under exchangeability [\hyperlink{bib.ref1}{1}, \hyperlink{bib.ref2}{2}]. For classification, the conformity score controls the coverage/efficiency trade-off: LAC/THR tends to produce small sets [\hyperlink{bib.ref3}{3}], APS uses cumulative sorted probability mass [\hyperlink{bib.ref4}{4}], and RAPS regularizes the APS tail for smaller sets [\hyperlink{bib.ref5}{5}]. Per-class or Mondrian calibration [\hyperlink{bib.ref7}{7}] can provide label-conditional guarantees when enough calibration data are available, while clustered conformal [\hyperlink{bib.ref6}{6}] pools classes to reduce the per-class sample burden.

\textbf{Conformal under distribution shift.} Weighted conformal prediction and related likelihood-ratio methods address covariate shift when it is known or estimable [\hyperlink{bib.ref8}{8}]; other methods adapt online [\hyperlink{bib.ref9}{9}] or without target labels [\hyperlink{bib.ref10}{10}]. Recent papers also study conformal prediction under shift, including WQLCP [\hyperlink{bib.ref13}{13}], group-partitioned conformal [\hyperlink{bib.ref29}{29}], and audited conformal prediction under unknown shift [\hyperlink{bib.ref31}{31}]. Those works focus on trained visual classifiers and marginal or group coverage. Our focus is frozen zero-shot VLMs, score geometry, and the worst-class class-conditional tail.

\textbf{OOD generalization and calibration.} A parallel non-conformal literature studies accuracy and confidence calibration under distribution shift, including pretrained-model selection for OOD generalization and calibration [\hyperlink{bib.ref44}{44}], where larger models and stronger pretraining corpora improve average accuracy and calibration out of distribution. Our scale and corpus results (Sec.~4.7) connect to this line: the average-level gains it documents do not carry over to worst-class conformal coverage, which remains collapsed across the tested scales and corpora.

\textbf{Conformal prediction for zero-shot VLMs.} The closest general-domain repair is Conf-OT [\hyperlink{bib.ref11}{11}], which uses transductive optimal transport to realign calibration and query scores and improve set efficiency under drift, on CLIP/MetaCLIP [\hyperlink{bib.ref35}{35}] with a fixed label space. Fillioux et al.~[\hyperlink{bib.ref12}{12}] benchmark conformal scores across many foundation models and report that APS is relatively robust in terms of marginal coverage.

Our results are complementary. We reproduce the marginal behavior these works emphasize, but show that marginal and average metrics do not capture worst-class tail safety. In our audit, class-conditional worst-class coverage can collapse to $\approx$0 beneath an acceptable marginal number (Sec.~4.2), and the transductive repair that restores marginal coverage leaves that tail unresolved (Sec.~4.4). Our audit combines these axes: CLIP/OpenCLIP/SigLIP family span, sigmoid-vs-softmax score geometry, corruption severity, vocabulary expansion, and worst-tail repair analysis. Empirical CP under shift on standard ViT and ConvNeXt classifiers reports the same pattern, with marginal coverage holding while class-conditional coverage is violated on ImageNet variants [\hyperlink{bib.ref30}{30}], corroborating Sec.~4.2 on a different model class.

\textbf{Adjacent VLM repair methods.} Recent medical-VLM work studies few-shot or training-based conformal adaptation, including SCA-T [\hyperlink{bib.ref32}{32}] and CaCT [\hyperlink{bib.ref33}{33}]. These methods require model access, target adaptation, or a medical-domain setting. We instead study a frozen, black-box, no-target-label deployment regime, where the available interventions are limited to source calibration, unlabeled-target transduction, selective prediction, or target-labeled oracle recalibration.

\textbf{Class-conditional and open-set conformal prediction.} Class-wise, group-wise, long-tail, and open-set conformal methods directly target stronger notions of coverage or new-label settings [\hyperlink{bib.ref23}{23}, \hyperlink{bib.ref24}{24}, \hyperlink{bib.ref25}{25}, \hyperlink{bib.ref26}{26}, \hyperlink{bib.ref27}{27}, \hyperlink{bib.ref28}{28}], and learned conditional-coverage diagnostics such as ERT [\hyperlink{bib.ref34}{34}] test conditional validity directly; our simulated null band is a lighter-weight alternative tailored to the worst-of-$K$ statistic. These are complementary to our audit. We do not solve open-set conformal prediction; instead, we test how standard VLM conformal scores behave as the vocabulary expands (Sec.~S8). Similarly, we do not propose a new class-wise repair; we evaluate whether standard deployable repairs already provide worst-class tail safety under shift.

\pdfbookmark[0]{3. Audit protocol}{mainsec.3}
\section{Audit protocol}
\textbf{Models and scores.} We evaluate three frozen zero-shot VLM families at matched ViT-B/16 scale: CLIP [\hyperlink{bib.ref14}{14}], OpenCLIP [\hyperlink{bib.ref15}{15}], and SigLIP [\hyperlink{bib.ref16}{16}], pretrained on OpenAI WIT, LAION-2B [\hyperlink{bib.ref42}{42}], and WebLI [\hyperlink{bib.ref43}{43}] respectively. We cache per-class logits for each model and dataset, so conformal analyses and bootstraps run from fixed logits. CLIP and OpenCLIP provide softmax probabilities; SigLIP provides independent per-class sigmoid scores, so we report both native SigLIP scores and softmax-normalized SigLIP logits when using probability-mass scores such as APS and RAPS. Unless otherwise stated, SigLIP results in cross-family comparisons use the softmax-normalized scores. We use each model's released prompt protocol; a matched single-prompt ablation (Appendix~E) leaves the cross-family set-size gap unchanged.

\textbf{Conformity scores.} For per-class scores $p_y$, we use three standard non-conformity scores [\hyperlink{bib.ref3}{3}, \hyperlink{bib.ref4}{4}, \hyperlink{bib.ref5}{5}]:

\[s_y^{\mathrm{LAC}}=1-p_y,\quad s_y^{\mathrm{APS}}=\sum_{j:\,p_j\ge p_y}p_j,\quad s_y^{\mathrm{RAPS}}=s_y^{\mathrm{APS}}+\lambda\max(0,\mathrm{rank}(y)-k_{\mathrm{reg}}).\]
APS uses the cumulative probability mass above the true label, and RAPS regularizes that tail for smaller sets; unless stated, $k_{\mathrm{reg}}{=}1$ and $\lambda{=}0.01$ (swept in Appendix~G). The conformal threshold $\hat q$ is the $k$-th smallest calibration score with $k=\lceil(n{+}1)(1-\alpha)\rceil$, which gives finite-sample marginal coverage under exchangeability. APS and RAPS have their usual interpretation only when the scores form a probability simplex. Native SigLIP sigmoids remain valid deterministic inputs to a conformal score, but they do not form a probability simplex; APS therefore loses its cumulative-mass interpretation and yields near-trivial sets (Sec.~S1). Cross-family APS/RAPS comparisons use softmax-normalized SigLIP logits.

\textbf{Conformal protocol.} We split the 50k-image ImageNet-val [\hyperlink{bib.ref39}{39}] into equal calibration and source-test folds, fit thresholds at $\alpha$=0.10, and evaluate the ID test fold and shifted datasets without recalibration; we also report a target-recalibrated oracle as a non-deployable upper bound. Each score (LAC, APS, RAPS) is run with both marginal and class-conditional (Mondrian) calibration. ImageNet-R and ImageNet-A are evaluated under a restricted-200 protocol: we calibrate and test only on the corresponding 200 ImageNet classes, rescoring the classifier over that label space, so their set sizes are not directly comparable to the 1000-way results. Results are averaged over 10 random splits with bootstrap 95\% confidence intervals ($\alpha$=0.05 in Appendix~B gives the same qualitative conclusions). The reported intervals quantify variability over calibration/test partitions of the same benchmark, not independent-dataset sampling.

\textbf{Metrics.} We report marginal coverage and its gap to $1{-}\alpha$, average prediction-set size (efficiency), and per-class coverage $\mathrm{Cov}(y)$ over classes with at least five evaluation examples. The per-class distribution is summarized by the minimum class-conditional coverage, MCCC $=\min_y \mathrm{Cov}(y)$, the 5th and 10th percentiles (p5/p10), the below-null and below-nominal fractions (classes below the finite-sample null floor and below the target $1{-}\alpha$), and the mean per-class under-coverage, CCV gap $=\mathrm{mean}_y\max(0,(1{-}\alpha)-\mathrm{Cov}(y))$. Because MCCC is a minimum over many classes and is dominated by worst-of-$K$ sampling noise, we compare it against a simulated null-MCCC band: the MCCC distribution a perfectly valid 0.90 predictor would produce given the observed per-class counts, so a collapse is only claimed when the observed value falls below that band. We lead with the percentiles and below-null/below-nominal fractions, which are more robust than MCCC alone. Exact distribution-free coverage conditional on the full input is impossible without further assumptions [\hyperlink{bib.ref38}{38}], so we audit the achievable label-conditional and tail diagnostics rather than claim full conditional coverage. Coverage and set size are always reported jointly: a method counts as an improvement only if it holds coverage and shrinks sets.

\textbf{Engine.} The conformal engine is validated by exact ID coverage, analytic Vovk-band checks, and synthetic-logit unit tests; all figures and tables are computed from cached logits and fixed seeds for determinism.

\pdfbookmark[0]{4. Experiments and results}{mainsec.4}
\section{Experiments \& results}
All experiments follow the Sec.~3 protocol: ImageNet-val calibration at $\alpha$=0.10, no target recalibration except the oracle, 10-split bootstrap 95\% CIs, SigLIP softmax-normalized unless stated, and ImageNet-R/A under the restricted-200 protocol. We evaluate on ImageNet-V2 [\hyperlink{bib.ref17}{17}], ImageNet-Sketch [\hyperlink{bib.ref20}{20}], ImageNet-R [\hyperlink{bib.ref18}{18}], ImageNet-A [\hyperlink{bib.ref19}{19}], ImageNet-C [\hyperlink{bib.ref21}{21}], and the non-ImageNet benchmarks Stanford Cars [\hyperlink{bib.ref40}{40}] and Food-101 [\hyperlink{bib.ref41}{41}].

\subsection{Marginal coverage is family- and shift-dependent}
Before examining the class tail, we first ask whether marginal behavior itself is stable across VLM families and shift types. APS-marginal coverage at the 0.90 target undercovers mildly on V2 (CLIP 0.888, OpenCLIP 0.871, SigLIP 0.877) and Sketch (0.861/0.860/0.864), stays near-valid on R under the restricted-200 protocol (0.901/0.875/0.905), and collapses on A, where the drop is strongly family-dependent: CLIP/OpenCLIP/SigLIP reach 0.817/0.649/0.739. LAC under-covers far more everywhere (e.g. Sketch 0.77--0.84, A 0.29--0.45) and RAPS sits between. No score is uniformly safe, and the family ordering flips by shift type: OpenCLIP (LAION) is most robust on renditions (R) but least on adversarial (A), where CLIP is most robust. This complicates the reading that APS least violates marginal coverage [\hyperlink{bib.ref12}{12}] once families and severities widen beyond CLIP and MetaCLIP. The average behavior is therefore already family-dependent, which motivates a stricter question: can a passable marginal number conceal failures for individual classes?

\subsection{Class-conditional tail coverage collapses below the finite-sample null under shift}
\begin{figure}[t]\centering
\includegraphics[width=0.88\linewidth]{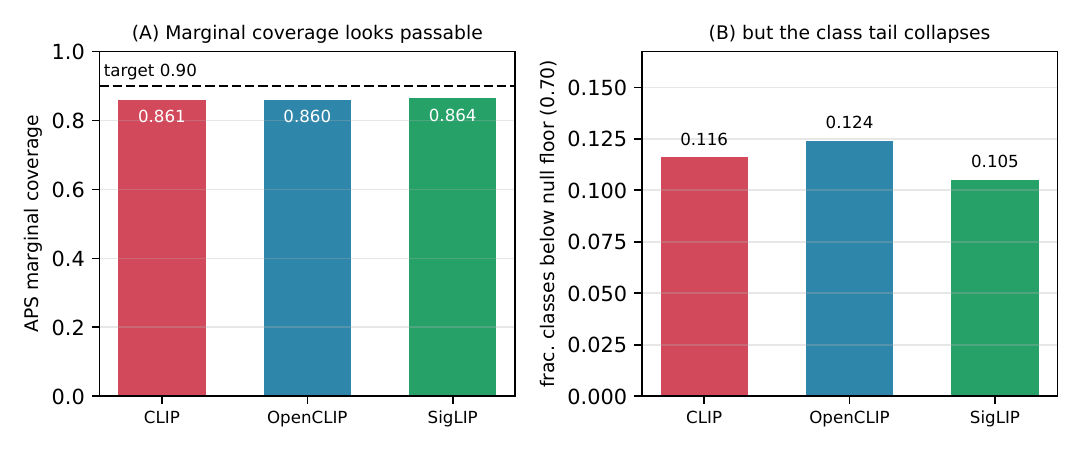}
\caption{ImageNet-Sketch, APS calibrated on ImageNet-val. Marginal coverage remains near 0.86 for all three families (A; target 0.90), while 10--12\% of classes fall below the finite-sample null floor of 0.70 (B) and the worst class receives zero coverage.}\label{fig:thesis}
\end{figure}

Figure 1 answers this question on ImageNet-Sketch: marginal coverage remains relatively high (about 0.86) across families, while a nontrivial lower tail of classes falls below the finite-sample reference and the worst class receives no coverage. We compare per-class coverage against a simulated null band: the coverage distribution a perfectly valid 0.90 predictor would yield from per-class sampling noise alone given the observed counts (at 50 images/class, essentially no class falls below 0.70). On the dense sets where this is unambiguous (ImageNet-Sketch at 50/class and ImageNet-R at 150/class) the collapse is real rather than an outlier, and we report percentiles to avoid over-reading the worst-of-$K$ minimum: it is widespread on Sketch and milder on R, whose restricted-200 below-floor fraction is only 1--7\% against Sketch's 10--12\%. On Sketch (APS-marginal), 10--12\% of classes fall below the 0.70 null floor and 41--44\% below the 0.90 target for every family, the bottom 5\% of classes sit at 0.52--0.57 coverage (p10 0.65--0.69), and the worst classes reach 0\%, whereas a valid predictor would put near 0\% below the floor (Fig.~\ref{fig:thesis}). All estimates carry tight bootstrap 95\% CIs over 10 splits. The mean per-class under-coverage (CCV gap) grows 2--10$\times$ under shift.

Class-conditional (Mondrian) calibration substantially lifts the in-distribution tail, raising APS worst-class coverage from about 0.20 to 0.52--0.55 on the ID fold (into the finite-sample regime, not to nominal), but it does not transfer under shift: the below-floor fraction is essentially unchanged on Sketch and R, because source per-class thresholds do not carry over to the target. Only the non-deployable oracle (target-recalibrated) substantially lifts it, to a finite-sample 0.75 that is still sub-nominal. Marginal calibration is not class-conditional safety, and the standard class-conditional fix breaks under distribution shift. The next question is why the failing classes and the thresholds they require change under shift.

\subsection{The class-tail failure is accuracy-aligned and is not predicted by the tested source-domain diagnostics}
\begin{figure}[t]\centering
\includegraphics[width=0.88\linewidth]{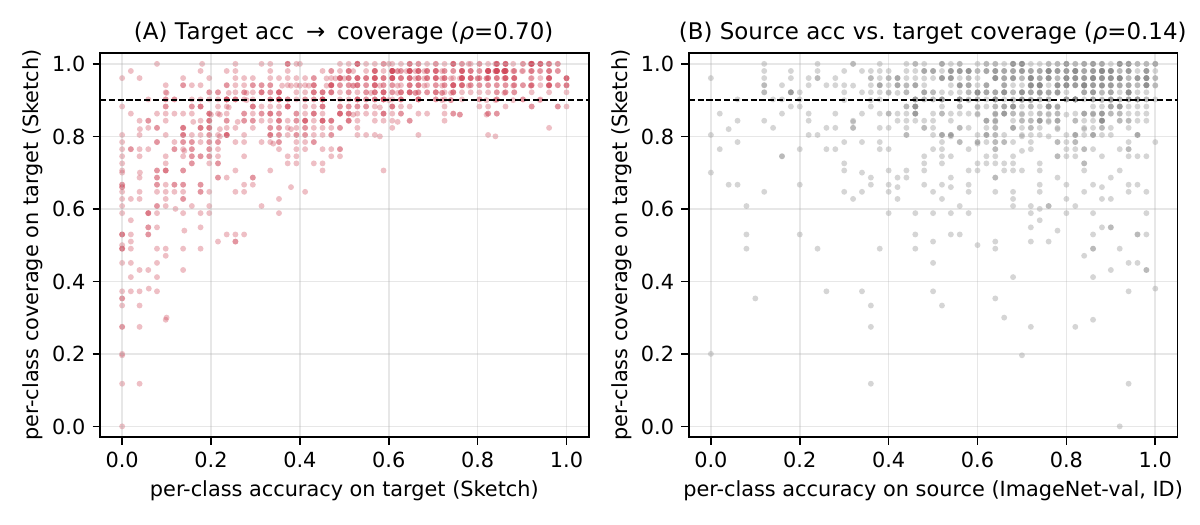}
\caption{Mechanism (CLIP, Sketch): per-class coverage tracks per-class target accuracy ($\rho{=}0.70$, left) but is essentially uncorrelated with per-class source accuracy ($\rho{=}0.14$, right), i.e.\ accuracy-driven yet source-invisible (Sec.~4.3).}\label{fig:mechscatter}
\end{figure}

Probing which classes collapse (per-class Spearman on ImageNet-Sketch, 50/class), per-class coverage under shift is strongly associated with per-class target accuracy ($\rho$=0.68--0.72 over about 1000 classes; Fig.~\ref{fig:mechscatter}(A)): coverage tends to fail where accuracy fails, class by class. On the collapsed classes the true label's median rank is 5--9 against 1 for covered classes, an accuracy and representation failure rather than a calibration near-miss. The collapse is partly shared across families and partly model-specific (per-family coverage $\rho$=0.53--0.66, about 30\% of worst classes shared).

A correctness decomposition shows why the marginal number stays passable: the marginal (0.86) is carried by the correct examples (true class in the set 0.97--0.99 of the time) against only 0.65--0.75 when the top-1 is wrong, so the worst classes collapse because incorrect examples dominate them. The accuracy link does not make the result trivial: marginal conformal coverage hides these class-level recall failures behind an averaged number, and the at-risk classes are not identified by the source-domain diagnostics we test. How much of the collapse is specific to the conformal score or calibration step, rather than inherited from the underlying zero-shot representation, cannot be fully separated in this audit: the accuracy alignment points to a large representation-driven component, while the score dependence of the below-null fraction (Sec.~4.5) shows the score choice modulates it.

None of the source-domain diagnostics we test predicts it: in-distribution per-class coverage, accuracy, confidence, and true-class probability all correlate only $\rho$=0.06--0.19 with which classes under-cover on the shift set (Fig.~\ref{fig:mechscatter}(B)). This source invisibility has a principled explanation: coverage is a true-class recall property, whereas the source and label-free target diagnostics available here primarily expose predicted-class structure, so a deployer cannot flag the at-risk classes without target labels. This is why source-calibrated Mondrian (Sec.~4.2) and Conf-OT's average-conditional repair (Sec.~4.4) leave the tail unfixed (qualitative class examples in the supplementary).

\begin{figure}[t]\centering
\includegraphics[width=0.88\linewidth]{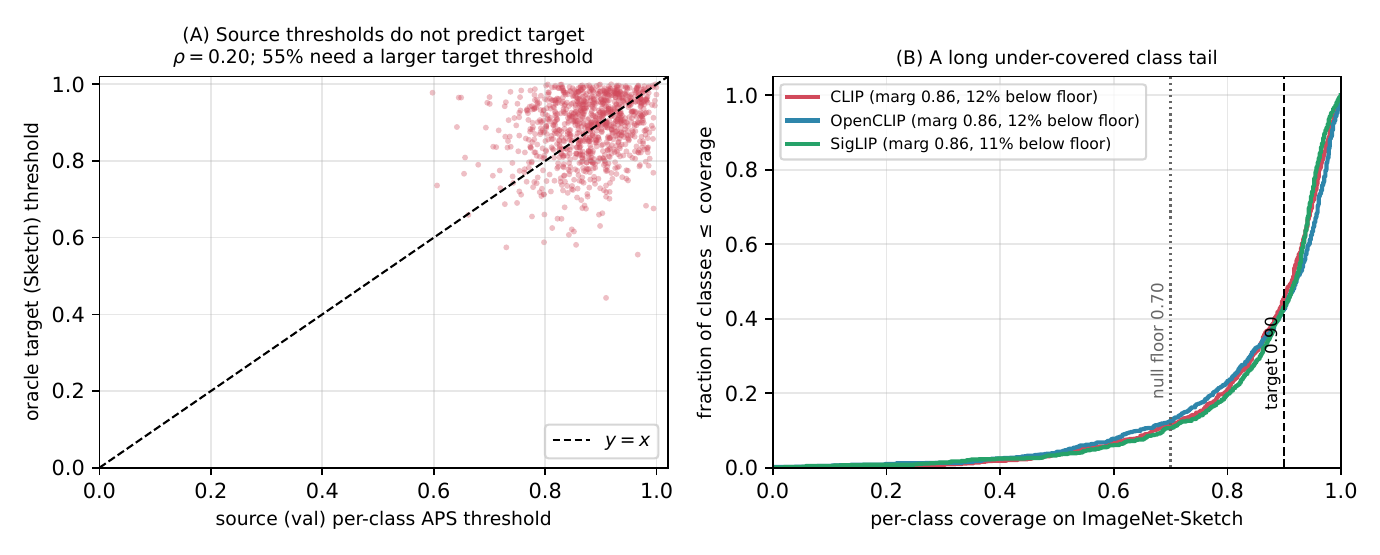}
\caption{Why source-side class calibration does not transfer (CLIP, ImageNet-Sketch). (a) Per-class APS thresholds learned on ImageNet-val weakly predict the oracle thresholds required on Sketch (Spearman $\rho{=}0.20$; 55\% of classes need a larger target threshold, points above $y{=}x$). (b) ECDF of per-class coverage: marginal coverage is near 0.86 for all three families, yet a long left tail falls below the 0.70 null floor (10--12\%) and the 0.90 target. Together the panels explain why source-Mondrian improves the in-distribution tail but fails under shift (Sec.~4.2--4.3).}\label{fig:twoviews}
\end{figure}

The source-side failure is also visible at the threshold level (Fig.~\ref{fig:twoviews}): the per-class thresholds Mondrian learns on val are nearly uncorrelated with those an oracle requires on Sketch (Spearman $\rho$=0.09--0.20 across families; about 55\% of classes need a larger target threshold), and the per-class coverage distribution has a broad under-covered left tail rather than a single noisy minimum. The shift reshuffles the entire per-class landscape, coverage and thresholds alike, which source-side calibration cannot anticipate. Increasing the marginal target does not remove the collapse: raising it to 0.999 lifts CLIP's worst class only to 0.31, at near-trivial sets of 453/1000 classes (Appendix~B), so it is not a too-loose-$\alpha$ artifact.

\subsection{The closest transductive repair (Conf-OT) restores marginal coverage and improves average behavior, but not the worst-class tail}
Given this mechanism, we test whether deployable repair methods recover the tail. Source-side thresholds do not transfer (Sec.~4.2), so the natural first candidate is an unlabeled-target transductive repair. We reimplement Conf-OT [\hyperlink{bib.ref11}{11}], a transductive Sinkhorn optimal-transport realignment of the combined calibration and query scores toward the calibration class prior, and confirm that our clean-room reimplementation reproduces its in-distribution behavior: at valid coverage it shrinks APS sets by 11--17\% (CLIP 11.6$\rightarrow$9.6), approaching the reported 20\% (our version is slightly conservative; Appendix~A). To rule out a hyperparameter artifact, we sweep the Sinkhorn settings (temperature $\tau\in\{0.5,1,2\}$, iterations $\in\{1,3,5,10\}$): marginal coverage (0.84--0.91) and set size (15--33) vary widely, but worst-class coverage on Sketch stays $\leq$0.02 in every setting (Appendix~A).

\begin{figure}[t]\centering
\includegraphics[width=0.88\linewidth]{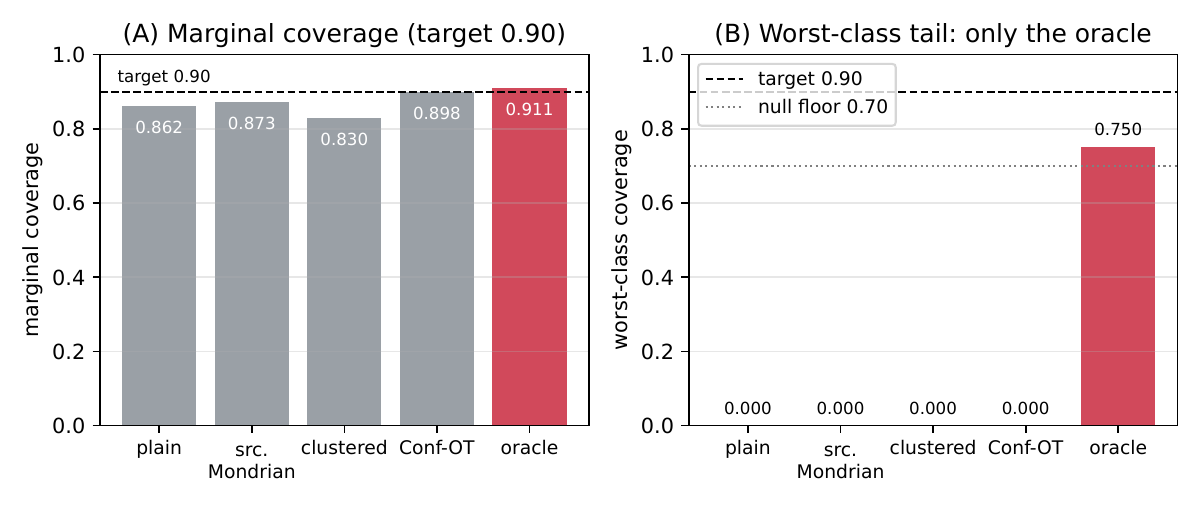}
\caption{Deployable repairs on ImageNet-Sketch (CLIP) improve different average metrics, and Conf-OT restores marginal coverage toward 0.90 (left), but none restores the worst-class tail (right, all at 0); only the target-label oracle substantially lifts the tail (Sec.~4.4).}\label{fig:repaircompare}
\end{figure}

Under shift, Conf-OT does what it is designed to do: on mild shift (V2) it shrinks APS sets about 12\% (CLIP 15.6$\rightarrow$13.6); on stronger shift (Sketch) it bridges the domain gap, restoring marginal coverage from about 0.86 back to 0.89--0.90 for all three families and roughly halving the mean per-class under-coverage (CLIP CCV gap 0.068$\rightarrow$0.037). Yet worst-class coverage remains 0.00 on Sketch for every family even after Conf-OT: it succeeds on the marginal and average-conditional metrics it targets, but those leave the safety-relevant worst-class tail unresolved (Fig.~\ref{fig:repaircompare}). Taken together with Sec.~4.2 and the oracle (which needs target labels), average metrics do not capture tail safety: among the deployable repairs we test (plain split-conformal, class-conditional calibration, clustered conformal, and transductive Conf-OT), none restores worst-class coverage under shift; only target-side class-conditional recalibration substantially lifts it, and even that is label-expensive and finite-sample sub-nominal (Sec.~S4).

\subsection{Score geometry and efficiency}
Coverage is only one deployment criterion: a predictor that achieves coverage only through near-full sets has limited operational value, so we also compare the efficiency and score geometry of the three families. Under the same source-calibrated $\alpha$=0.10 protocol, with comparable (not exactly matched) marginal coverage, APS-marginal average set size diverges sharply and consistently by 2--3$\times$: CLIP is the least efficient and SigLIP(softmax) the most, with the gap widening as severity grows (non-overlapping CIs throughout).

\begin{table}[!ht]\centering\small\setlength{\tabcolsep}{6pt}
\caption{APS-marginal mean set size by family across shifts ($\alpha$=0.10). CLIP is consistently least efficient; softmax-normalized SigLIP is most efficient.}\label{tab:4}
\resizebox{\ifdim\width>\linewidth \linewidth\else\width\fi}{!}{%
\begin{tabular}{lrrrr}
\toprule
family & ID & V2 & Sketch & R(200) \\
\midrule
CLIP & 11.5 & 15.8 & 26.7 & 5.9 \\
OpenCLIP & 5.5 & 7.4 & 14.1 & 2.7 \\
SigLIP(softmax) & 5.0 & 7.2 & 9.3 & 2.7 \\
\bottomrule
\end{tabular}}
\end{table}
CLIP pays a 2.1--2.9$\times$ larger set-size tax than the LAION/WebLI families at comparable marginal coverage. This is not an accuracy artifact: restricting to each family's correctly-classified images (matching accuracy at 100\%), CLIP still needs about 2.3$\times$ larger APS sets (7.25 vs OpenCLIP 3.18, SigLIP 2.86 on ID), so the divergence is consistent with the tested CLIP score geometry, a flatter softmax with more probability mass off the correct top class, independent of whether it is right. Nor is it a prompt-protocol artifact: re-extracting all three families with one matched single prompt (replacing CLIP/OpenCLIP's 80-template ensemble) leaves the ordering intact and in fact widens the Sketch gap (CLIP 2.84$\times$$\rightarrow$3.21$\times$ vs SigLIP), because the ensemble was sharpening CLIP/OpenCLIP and shrinking their sets, so the differing prompt protocol works in CLIP's favor (Appendix~E).

The gap is consistent with CLIP's flatter softmax (ID predictive entropy 1.27 nats vs 0.69/0.73), which inflates APS sets, but it is not a mere temperature artifact: no temperature buys both efficiency and class-conditional coverage, since sharpening shrinks sets only by collapsing the worst-class tail toward 0. The deployment implication concerns backbone choice: OpenCLIP and softmax-normalized SigLIP are substantially more efficient than CLIP in the tested settings, a contrast invisible to single-family studies. Together, the efficiency gap and the tail collapse suggest a family-dependent score-geometry trade-off: sharper scores buy smaller sets but can leave hard classes uncovered (full sweep in the supplementary).

\textbf{Which conformity score?} The efficiency gap above is one half of the score-geometry picture; the other is which non-conformity score to deploy. Averaging over three families and four shift sets ($\alpha$=0.10), the score choice trades coverage, efficiency, and robustness to SigLIP's score geometry:

\begin{table}[!ht]\centering\small\setlength{\tabcolsep}{6pt}
\caption{Conformity-score comparison averaged over three families and four natural-shift sets (V2/Sketch/R/A, $\alpha$=0.10): mean coverage, set size, CCV gap, and SigLIP native-sigmoid set size (V2/Sketch). RAPS is the efficiency--score-geometry compromise, not a tail-safety solution. ImageNet-R/A use a restricted-200 label space, so the mean set size is a within-protocol summary and is not directly comparable across the 1000-way and 200-way shifts.}\label{tab:5}
\resizebox{\ifdim\width>\linewidth \linewidth\else\width\fi}{!}{%
\begin{tabular}{lrrrr}
\toprule
score & mean coverage & mean set size & mean CCV gap & SigLIP-sigmoid set size (V2/Sketch) \\
\midrule
LAC/THR & 0.698 & 1.99 & 0.220 & 4 / 2 \\
APS & 0.842 & 9.57 & 0.081 & 721 / 940 \\
RAPS & 0.818 & 4.32 & 0.103 & 42 / 64 \\
\bottomrule
\end{tabular}}
\end{table}
LAC gives the smallest sets but under-covers severely under shift (mean coverage 0.70, 20 points below target). APS is the most reliable marginally but the least efficient, and it breaks down on SigLIP's native sigmoid (700--940-class sets). RAPS gives the best efficiency--score-geometry compromise: near-APS coverage at roughly half the set size and much smaller native-SigLIP sets; for SigLIP, APS and RAPS should be applied to softmax-normalized scores rather than native sigmoids. This is not a tail-safety solution: no score reaches nominal marginal coverage under strong shift (Sec.~4.1), and on Sketch RAPS worst-class coverage remains 0.00 for every family, with a below-null fraction in fact higher than APS (e.g. CLIP 0.20 vs 0.12), so the worst-class tail persists across the tested scores and remains the oracle's alone (Sec.~4.2, Sec.~4.4). The recommendation is stable to $k_{\text{reg}}$ within the tested grid, while $\lambda$ controls the coverage--efficiency trade-off; across the sweep RAPS still shrinks the APS set at every setting and preserves the native-SigLIP set-size reduction (Appendix~G).

\subsection{Putting it together: a deployment decision table}
The preceding results leave three deployment goals that require different information: restore marginal coverage, reduce set size on a retained subset, or recover the full-input class tail. Table~\ref{tab:1} summarizes the label-access trade-off on ImageNet-Sketch (CLIP; all three families in Appendix~J). Without target labels, no method restores the worst-class tail: plain, source-Mondrian, and clustered leave it at 0.00, and Conf-OT restores marginal coverage to about 0.90 but leaves the tail at 0.00. The oracle lifts the worst class to 0.75, but only with labeled target data: a budget sweep shows that even 15 labels per class (15k for ImageNet-1k) reach only about 0.53, and at the tested budgets of five or fewer labels per class the conservative per-class threshold is infinite, so validity holds only through full prediction sets (Sec.~S4). The one label-free method that recovers usable coverage is size-thresholded selective prediction, a conformal instance of selective classification [\hyperlink{bib.ref22}{22}]: deferring large-set inputs ($k{\le}5$) empirically recovers near-nominal retained coverage (0.88--0.90) on 47--74\% of inputs, empirical because selecting on set size changes the evaluated population. The appropriate method therefore depends on label access: Conf-OT for an unlabeled target batch, selective prediction for a retained subset, cross-shift transfer (Sec.~S3) given a labeled harder domain, and target recalibration only when target labels are available.

\begin{table}[!ht]\centering\small\setlength{\tabcolsep}{6pt}
\caption{Methods on ImageNet-Sketch (CLIP, APS, $\alpha$=0.10; all families in Appendix~J). Coverage is full-input marginal coverage for the all-input methods and empirical retained coverage for selective prediction; the latter is computed on a retained population and is not directly comparable with full-input worst-class coverage. Worst-class is the minimum class-conditional coverage. The rows target different objectives (marginal, class-conditional, or retained-subset coverage), so the table summarizes the label-access trade-off rather than ranking methods on one criterion.}\label{tab:1}
\resizebox{\ifdim\width>\linewidth \linewidth\else\width\fi}{!}{%
\begin{tabular}{lrrrrr}
\toprule
method & target information & evaluated population & coverage & worst-class cov & mean set size \\
\midrule
plain split-conformal & none & all inputs & 0.862 & 0.00 & 26.7 \\
source-Mondrian & none & all inputs & 0.873 & 0.00 & 45.3 \\
clustered conformal & none & all inputs & 0.830 & 0.00 & 29.6 \\
Conf-OT (transductive) & unlabeled target & all inputs & 0.898 & 0.00 & 27.9 \\
selective (k$\leq$5) & none & retained 47\% & 0.880 & N/A & 2.3 \\
oracle (target Mondrian) & labeled target & all inputs & 0.911 & 0.75 & 39.2 \\
\bottomrule
\end{tabular}}
\end{table}
\subsection{Scaling improves accuracy but not class-tail coverage}
\begin{figure}[t]\centering
\includegraphics[width=0.88\linewidth]{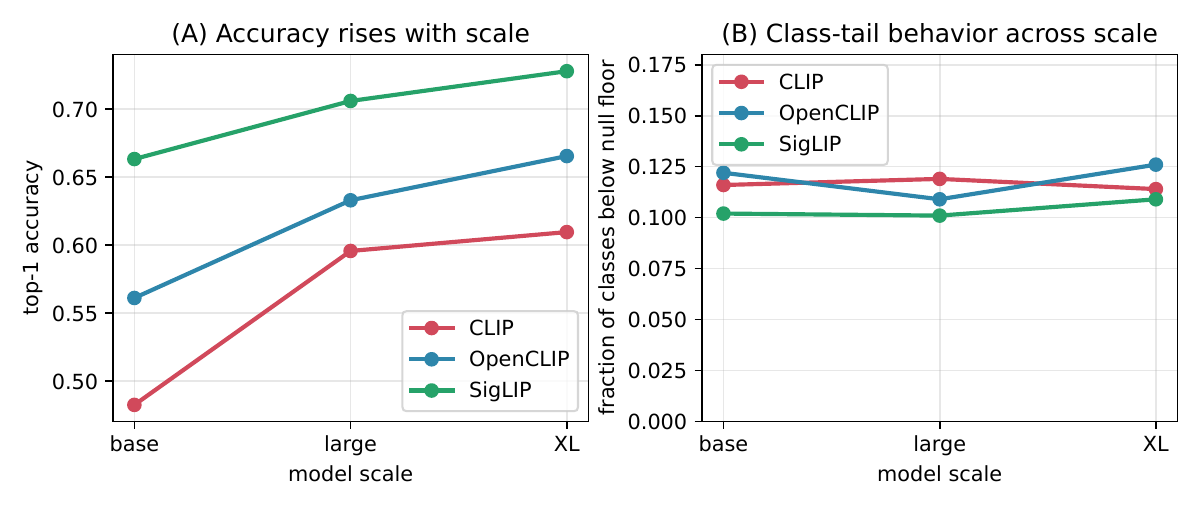}
\caption{Scale improves top-1 accuracy but leaves the class-tail failure nearly unchanged on ImageNet-Sketch: across CLIP, OpenCLIP, and SigLIP the below-null class fraction stays $\approx$10--13\% from base to XL.}\label{fig:scale}
\end{figure}

To test whether these effects are a small-backbone artifact, we replicate the audit across a base$\rightarrow$large$\rightarrow$XL scale ladder per family (to OpenCLIP ViT-H/14 and SigLIP-so400m) on ImageNet-Sketch. Scale sharply improves accuracy (CLIP 0.48$\rightarrow$0.61, OpenCLIP 0.56$\rightarrow$0.67, SigLIP 0.66$\rightarrow$0.73) and shrinks set size (CLIP 26.7$\rightarrow$15.2), but the coverage failures persist: marginal APS coverage stays 0.85--0.87 at every scale and the below-null fraction stays 0.10--0.13 (Fig.~\ref{fig:scale}). The strongest model we test, SigLIP-so400m (72.8\% Sketch accuracy), still leaves 11\% of classes below the finite-sample null floor, and the family efficiency gap (Sec.~4.5) persists at every tier. Larger VLMs improve average accuracy and efficiency, but a lower tail of difficult classes remains, so the fraction below the finite-sample coverage floor changes little (Sec.~4.3). The full ladder table is in Appendix~C.

\textbf{Robustness consolidation.} We vary each axis in turn to test whether our conclusions are an artifact of one checkpoint, corpus, prompt, $\alpha$, or dataset. Across the tested axes the corresponding conclusions persist: the class-tail result is supported by scale (the below-null tail holds at 10--13\% of classes across the ladder), pretraining corpus (all six reproduce the Sketch tail collapse), $\alpha$=0.05 (7--9\% below null), and the non-ImageNet analyses, while the matched-single-prompt control supports the cross-family efficiency gap (the set-size gap persists). Easy Food-101 ID is a near-negative control with negligible below-null breadth (Sec.~4.3), while Stanford Cars and shifted variants collapse where per-class accuracy is heterogeneous or shifted. Per-axis numbers are in the supplementary (Appendix~B, C, E, M, and N).

\pdfbookmark[0]{5. Limitations}{mainsec.5}
\section{Limitations}
\textbf{Statistical resolution.} ImageNet-val gives only about 50 calibration images per class, which is tight for per-class (Mondrian) thresholds; we address this on the evaluation side by judging MCCC against a simulated null band and by leaning the main finding on the densest sets (Sketch 50/class, R 150/class), where the collapse is unambiguous, while sparser sets (V2, A) carry the band only as context. The below-null fraction is a descriptive measure of how broadly the tail failure extends, not a calibrated per-class error probability.

\textbf{Scope of generalization.} We test three checkpoints per family across a base$\rightarrow$XL ladder, and the findings persist across it (Sec.~4.7), but each tier is a single checkpoint, so family claims are over the tested representatives. All tested encoders are ViT-based; convolutional backbones (e.g.\ ResNet-50/101) are untested here, although the same marginal-versus-class-conditional pattern has been reported for both ViT and ConvNeXt classifiers [\hyperlink{bib.ref30}{30}]. The open-vocabulary protocol is a constructed exchangeability violation, specified in advance and reported as a diagnostic rather than a full open-set treatment. Most experiments are ImageNet-scale recognition and its standard shift variants (V2/R/A/Sketch/-C) plus Cars/Food replication, not medical, satellite, autonomous-driving, or broad open-vocabulary deployments, and the ImageNet-A score-shift mechanism may be specific to A's adversarial construction. Marginal-coverage degradation partly tracks base accuracy, but the family efficiency gap survives an accuracy control: on correctly-classified images CLIP still needs about 2.3$\times$ larger sets (Sec.~4.5), so that effect is not an accuracy artifact.

\textbf{Scope of repair and safety.} Our statement that no deployable repair restores the tail is scoped to the full-input, no-target-label or unlabeled-target methods we test (plain, source-Mondrian, clustered conformal, transductive Conf-OT); only the labeled-target oracle lifts the tail, and only sub-nominally. Retraining and adaptation methods (CaCT, SCA-T) and the distinct open-set validity problem are out of scope for a frozen audit, and weighted CP and WQLCP [\hyperlink{bib.ref8}{8}, \hyperlink{bib.ref13}{13}] primarily target marginal coverage under covariate shift; whether a particular density-ratio estimator also improves class-tail behavior is left for future work. Finally, because we measure coverage and set size rather than downstream task harm, the safety framing concerns the statistical safety promise under shift rather than a measured task-level failure; and size-thresholded selective prediction recovers retained coverage only empirically, since the retained population is selected using the model's own output, so it is not a new finite-sample conformal guarantee. The source-invisibility conclusion is limited to the diagnostics we test; it is not an impossibility result for all nonlinear or multivariate source-derived predictors.

\pdfbookmark[0]{6. Conclusion}{mainsec.6}
\section{Conclusion}
We audit split-conformal prediction for CLIP, OpenCLIP, and SigLIP under visual and vocabulary shift. A relatively high marginal number can conceal a severe class-conditional tail failure that is aligned with target class difficulty but not predicted by the source-domain diagnostics evaluated here. Source-Mondrian, clustered conformal, and Conf-OT improve source, marginal, or average behavior without recovering the worst-class tail; target-side class calibration lifts the tail only at substantial label and set-size cost. Score and backbone choice materially affect efficiency, and size-based deferral offers an empirical retained-population trade-off rather than full-input tail safety. Marginal conformal coverage should therefore be interpreted as an average reliability statistic, not as a guarantee for the class tail.

\clearpage
\setcounter{table}{0}\setcounter{figure}{0}
\renewcommand{\thetable}{S\arabic{table}}
\renewcommand{\thefigure}{S\arabic{figure}}
\pdfbookmark[0]{Supplementary Material}{suppmat}
\begin{center}
{\Large\bfseries Supplementary Material}\\[4pt]
{\bfseries Does Marginal Coverage Guarantee Class-Conditional Safety\\ for Zero-Shot VLMs Under Shift?}
\end{center}
\medskip

\textbf{Supplement organization.} The main paper establishes that marginal coverage can hide a class-tail failure; this supplement tests the most likely alternative explanations and deployment responses. Part I asks whether the result is caused by score geometry or sparse class-wise calibration, and whether source, proxy, or target calibration repairs it (S1--S4). Part II studies how the failure evolves with corruption severity and prediction-set size, then evaluates size-based deferral and vocabulary expansion (S5--S8). Part III tests generalization across model scale, pretraining corpus, and non-ImageNet label spaces (S9--S11). The appendices provide the full protocols, configurations, tables, and the Conf-OT reproduction; a positioning table relative to the closest conformal-VLM and shift-conformal work is in Appendix~Q. The claim map below points each supporting result to its section. Citations [N] follow the numbered reference list in the main paper.

\begin{table}[!ht]\centering\small\setlength{\tabcolsep}{6pt}
\caption{Claim map: each supporting result and the section that establishes it.}\label{tab:3}
\begin{tabularx}{\linewidth}{X p{0.22\linewidth}}
\toprule
claim & section \\
\midrule
score compatibility (native vs softmax SigLIP) & S1 \\
additional repair baselines (clustering, pseudo-label) & S2 \\
proxy- and target-label calibration & S3--S4 \\
severity and size conditioning & S5--S6 \\
size-based deferral & S7 \\
vocabulary expansion & S8 \\
scale, corpus, and non-ImageNet robustness & S9--S11 \\
\bottomrule
\end{tabularx}
\end{table}
\supppart{Is the failure an artifact, and can calibration repair it?}
\suppsection{Native SigLIP scores remove APS's probability-mass interpretation}
SigLIP's per-class sigmoid scores are formally usable as deterministic conformal scores, but they are not a probability simplex, so APS's cumulative-mass construction is no longer meaningful and inflates the prediction set to 568 (ID) / 721 (V2) / 940 (Sketch) classes vs $\approx$5--27 for the softmax families; its apparent "overcoverage" (e.g. 0.94 on V2) is this artifact, not reliability. RAPS's tail regularization mitigates it and LAC is unaffected. Softmax-normalizing SigLIP's logits collapses the APS set size $\approx$115$\times$ (568$\rightarrow$4.95 on ID) and makes SigLIP the most efficient family. We therefore use softmax-normalized SigLIP logits for all cross-family APS/RAPS comparisons. We next test whether additional class-conditional repairs recover the tail.

\suppsection{Additional repair baselines: clustering and pseudo-label Mondrian do not restore the tail}
Beyond the main-paper repairs, we test two further deployable baselines. Clustered conformal [6] pools sparse per-class data into similarity groups, the natural remedy if the collapse were a 50-images/class sparsity artifact. Pooling similar source classes does not improve the Sketch tail: worst-class coverage stays $\approx$0 and the below-null fraction does not decrease. The result does not support a sparsity-only explanation, since pooling source classes does not address the source-to-target rearrangement of class difficulty that main-paper Fig. 3 visualizes (Main Sec.~4.3).

Transductive pseudo-label Mondrian, the natural label-free per-class fix, pseudo-labels the target by argmax and sets per-class thresholds from it. It performs substantially worse than plain split-conformal: marginal coverage falls to 0.61--0.71 and 36--57\% of classes fall below the null floor (vs ~10--12\% for plain). This is consistent with the recall-side failure identified in the main paper: the pseudo-labels are wrong precisely for the hard, under-covered classes whose target thresholds need the most correction (Main Sec.~4.3).

Only target-label recalibration (the oracle) substantially lifts worst-class coverage, to $\approx$0.75, though it remains below nominal. Together with the Conf-OT result in the main paper, these baselines distinguish average repair from tail repair: methods that target marginal or average-class-conditional coverage leave the per-class tail collapse untouched.

\suppsection{Harder-proxy calibration is empirically conservative within a fixed shift chain}
Target labels are rarely available, so we ask whether calibrating on a different shift transfers. Within the shared 1000-class chain val$\rightarrow$V2$\rightarrow$Sketch (full matrices in Appendix~F), calibrating on the harder domain in the chain gives empirical coverage at or above nominal on the easier target (Sketch$\rightarrow$val 0.93, Sketch$\rightarrow$V2 0.906--0.919, V2$\rightarrow$val 0.91--0.93), while calibrating on a milder domain undercovers the harder target (val$\rightarrow$Sketch $\approx$0.86, V2$\rightarrow$Sketch $\approx$0.88); the structure is identical across families. This is specific to the tested chain and is not a general severity ordering, and it repairs only marginal coverage: the class-conditional tail still needs target-specific recalibration (Main Sec.~4.2/4.4). It is a diagnostic transfer pattern, not a general method for identifying a suitably hard proxy domain before deployment. (V2 is a re-collection shift and Sketch a domain/style shift; ImageNet-R/-A are excluded, their 200-class subsets sharing only 86/200 classes.)

\begin{figure}[ht]\centering
\includegraphics[width=\linewidth]{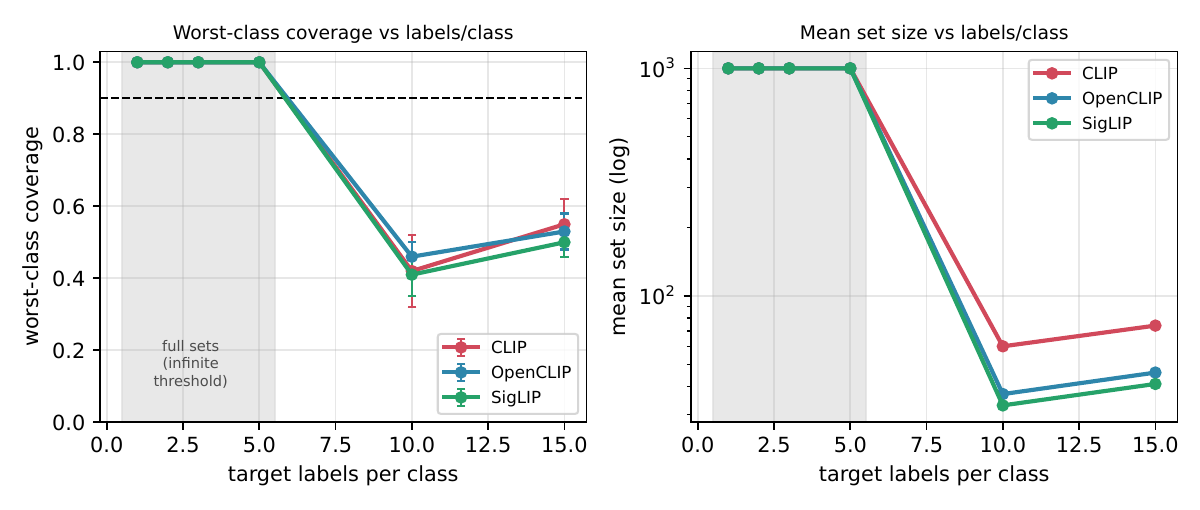}
\caption{Target-side Mondrian worst-class recovery (left) and mean set size (right, log scale) vs.\ per-class target-label budget (ImageNet-Sketch, APS; 5 seeds). At ${\le}5$ labels/class the conformal rank exceeds the per-class sample, so thresholds are infinite and sets are full (shaded region); finite-threshold recovery begins at ${\ge}10$ labels/class and stays sub-nominal (Main Sec.~4.6).}\label{fig:labelbudget}
\end{figure}

\suppsection{Target-side Mondrian calibration is label- and set-size-intensive}
The oracle substantially lifts the worst class, but at what labeling cost? We vary the per-class target-label budget $n$ used to estimate the per-class APS thresholds, using the same conservative conformal quantile as the engine (held-out, 5 seeds; Fig.~\ref{fig:labelbudget}). For very small per-class samples the conformal rank $\lceil(n{+}1)(1{-}\alpha)\rceil$ exceeds the sample count, so the threshold is infinite and validity is obtained only through near-trivial full prediction sets: at $n{\le}5$ labels/class, worst-class and marginal coverage are both 1.0 with mean set size 1000 (all classes). Once the thresholds become finite ($n{\ge}10$), worst-class coverage recovers slowly and stays sub-nominal: about 0.41--0.46 at 10 labels/class and 0.50--0.55 at 15, near-identical across families, while sets stay large (mean 33--74).

This is the Mondrian sample-complexity tax [7]: per-class CP turns the calibration budget into a per-class budget, so the worst-of-1000 minimum converges slowly, and 15 labels for every class (15k target labels) still leave the worst class near 0.53, far below the 0.90 target.

The same sweep on the milder ImageNet-R shift (restricted-200; Appendix~O) shows the same qualitative transition from trivial full sets to gradual, sub-nominal recovery (about 0.55 at 10 labels/class, 0.61 at 20). Because R differs from Sketch in label space and per-class count, we read this as a general per-class sample-complexity cost rather than a severity law. This motivates the label-free deployment responses (Conf-OT, selective prediction, cross-shift transfer; Main Sec.~4.6) examined next.

\begin{figure}[ht]\centering
\includegraphics[width=\linewidth]{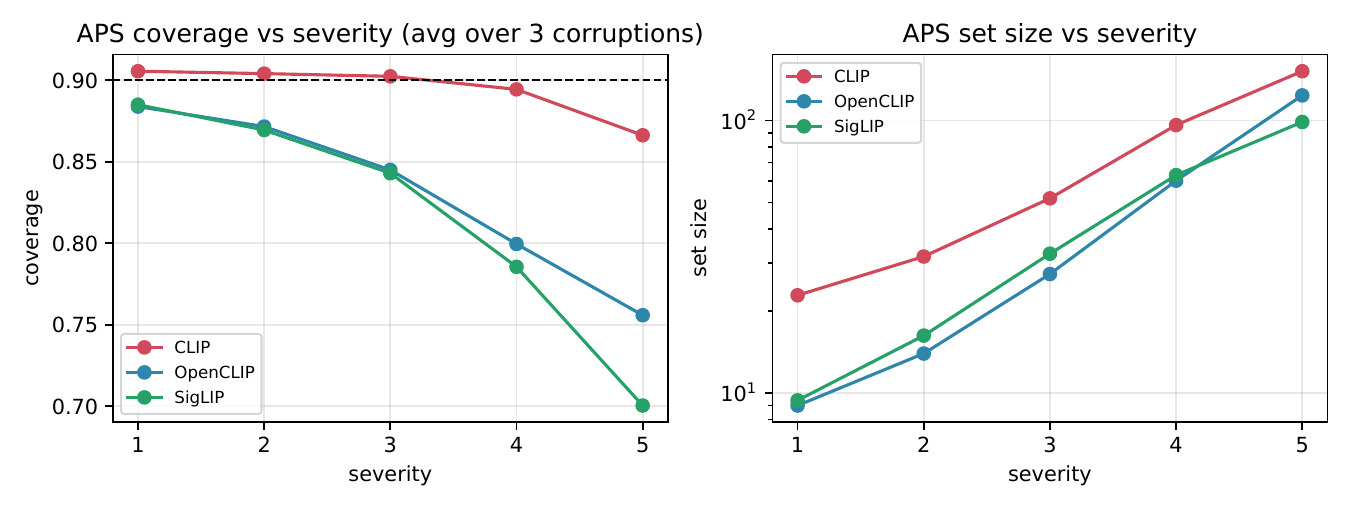}
\caption{APS marginal coverage and mean set size across three representative ImageNet-C corruptions (Gaussian noise, JPEG, motion blur), averaged at each severity 1--5. Coverage decreases and mean set size grows with severity for every family; the dashed line marks the 0.90 target (Main Sec.~4.1).}\label{fig:severity}
\end{figure}

\supppart{How does the failure evolve, and can abstention manage it?}
\suppsection{ImageNet-C severity increases set size while reducing marginal coverage (Fig.~\ref{fig:severity})}
We add a controlled severity axis using Gaussian noise, JPEG compression, and motion blur at severities 1$\rightarrow$5. Averaged across these three representative corruptions, APS marginal coverage decreases with severity for every family (CLIP 0.905$\rightarrow$0.866, OpenCLIP 0.884$\rightarrow$0.756, SigLIP 0.885$\rightarrow$0.700), sharper on noise (SigLIP 0.53 at gaussian-noise severity 5) and gentler on JPEG, with the family ordering identical across all three. Meanwhile the APS set size grows sharply (to 146--271 classes at gaussian-noise severity 5), so under strong corruption the predictor is simultaneously inefficient and empirically under-covered, and LAC coverage falls to 0.15--0.32. This complements the natural-shift audit with a controlled severity axis over three corruption types (noise/digital/blur), not the full ImageNet-C suite. We restrict this severity experiment to marginal metrics because the 5k subset is too sparse for stable class-tail localization; the dense class-tail analysis is reported in Main Sec.~4.2/4.3.

\begin{figure}[ht]\centering
\includegraphics[width=0.72\linewidth]{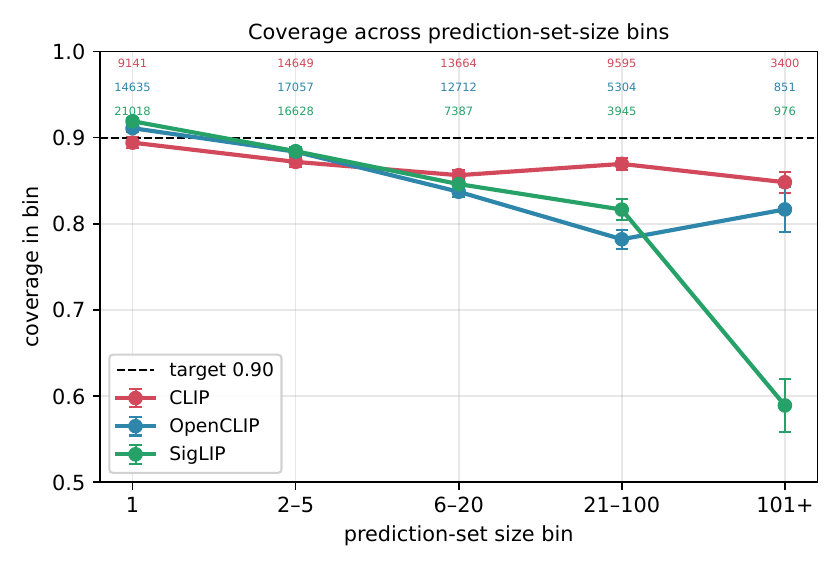}
\caption{Coverage across prediction-set-size bins on ImageNet-Sketch: the largest-set bins fall below the 0.90 target for every family, though the relationship is not strictly monotone. Error bars are 95\% binomial intervals and per-bin example counts are annotated for each family (in family colours); the largest-set bins hold relatively few examples (Main Sec.~4.6).}\label{fig:sizecov}
\end{figure}

\suppsection{Large prediction sets concentrate empirical undercoverage}
Binning ImageNet-Sketch test images by APS set size and measuring coverage within each bin (size-stratified coverage following [5]; Appendix~H, Fig.~\ref{fig:sizecov}), the largest bins (101+ classes) fall below the 0.90 target for every family (0.59--0.85 across families), while OpenCLIP and SigLIP singletons exceed the 0.90 target and CLIP singletons are near nominal at 0.894. The relationship is not strictly monotone and the largest bins hold few examples, but large sets consistently mark a less reliable regime, sharpest for SigLIP and mildest for CLIP (mirroring their set-size distributions, Main Sec.~4.5). This is a size-conditional analogue of the class-conditional collapse (Main Sec.~4.2) and motivates size-based abstention: coverage is lowest in the large-set regime, behind an acceptable marginal number.

\begin{figure}[ht]\centering
\includegraphics[width=\linewidth]{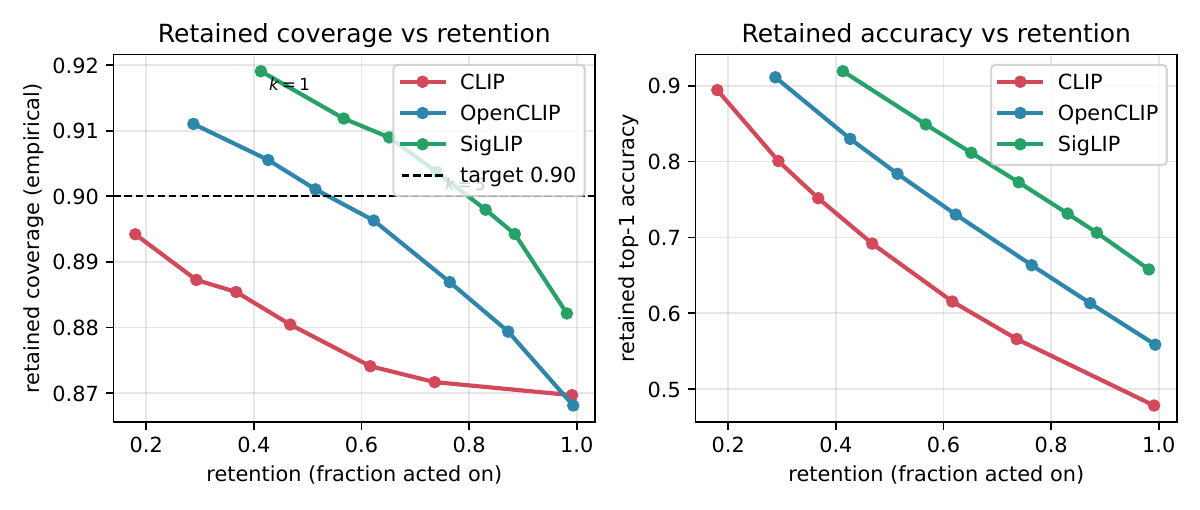}
\caption{Empirical retained coverage and top-1 accuracy under size-thresholded selection on ImageNet-Sketch, as a function of retention. Points correspond to discrete set-size caps $k$; the $k{=}1$ and $k{=}5$ operating points are annotated. Coverage here is empirical because selection changes the evaluated population (Main Sec.~4.6).}\label{fig:selective}
\end{figure}

\suppsection{Size-thresholded selective prediction recovers empirical retained coverage}
Because selecting on set size changes the evaluated population, the retained-coverage numbers here are empirical and are not finite-sample conformal guarantees. Set size is an actionable abstention signal (Sec.~S6): act on small sets, defer large ones. Capping the retained APS set size at $k$ on ImageNet-Sketch (Appendix~I) trades retention for reliability along a smooth, family-dependent frontier (Fig.~\ref{fig:selective}): retaining only singletons recovers near-nominal coverage at high accuracy on 18--41\% of inputs, and a looser cap $k{\le}5$ keeps SigLIP at 0.90 coverage / 0.77 top-1 on 74\% of inputs (full family frontier in the figure). SigLIP dominates and CLIP trails, matching the efficiency ranking (Main Sec.~4.5). Set size is on par with raw top-1 confidence as a ranking signal (within $\approx$0.01 coverage at matched retention; Appendix~L); its value is that it is the native conformal output and directly controls retained-set complexity.

Conf-OT and selective prediction compose: layering the size threshold on Conf-OT's transported codes (Main Sec.~4.4) lifts retained coverage a further $\approx$2--3 points (to 0.91--0.92 at $k{\le}5$), the best observed label-free trade-off, on 45--74\% of inputs with $\approx$2-class sets using only the unlabeled target batch (Appendix~I). This approach abstains on the hard tail rather than covering it; full-input class-tail safety remains unresolved.

\suppsection{Vocabulary-expansion stress test: LAC is sensitive, APS is stable}
We calibrate conformal on a vocabulary of $k_{\text{seen}}$ classes (val images re-scored $k_{\text{seen}}$-way), then deploy with the vocabulary expanded to all 1000 (novel text-defined distractors), keeping the true label in the seen set. LAC/THR coverage decreases with vocabulary size, falling from 0.90 to $\approx$0.78--0.80 after adding 750 distractors (tightly reproducible across 10 partitions, LAC penalty std $\leq$0.015; full dose-response in Appendix~P). APS is substantially more stable, with a maximum observed penalty of 0.037, consistent with most distractors ranking below the true class, which leaves APS's cumulative-mass-above-truth score largely unchanged while LAC's absolute-probability threshold is diluted by softmax renormalization. Target-side recalibration restores nominal coverage. This is a vocabulary-expansion stress test, not a full open-set conformal benchmark: the true label always remains in the seen vocabulary. We finally test whether the failures persist across model scale, pretraining corpus, and label space.

\begin{figure}[ht]\centering
\includegraphics[width=0.72\linewidth]{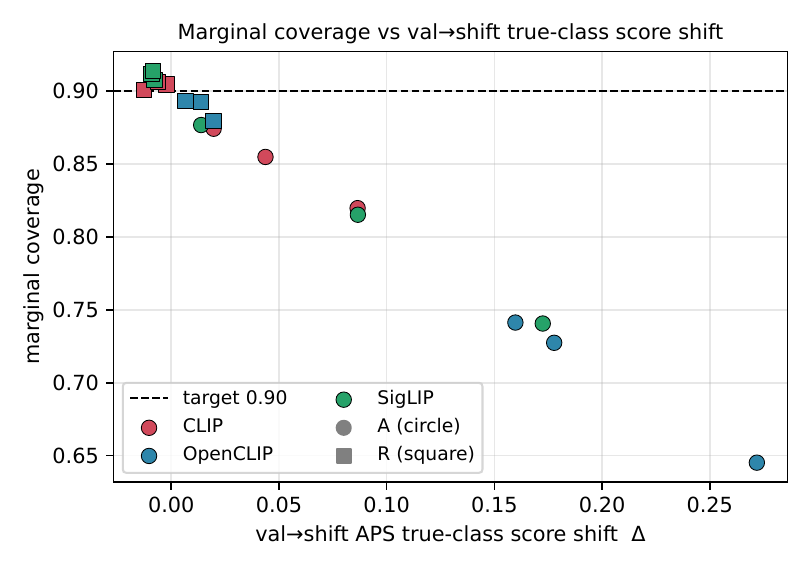}
\caption{Marginal coverage vs.\ the val$\rightarrow$shift true-class APS score shift $\Delta$, across the tested ImageNet-A (circles) and ImageNet-R (squares) checkpoints. A larger shift is associated with lower marginal coverage (Spearman $\rho{=}-0.97$ over the pooled points); the dashed line marks the 0.90 target (Main Sec.~4.7).}\label{fig:mechanism}
\end{figure}

\supppart{Does the conclusion generalize?}
\suppsection{ImageNet-A marginal undercoverage is associated with score shift}
Under natural-adversarial shift (ImageNet-A) the marginal collapse behaves oppositely to the Sketch conditional tail across the same ladder: marginal APS coverage rises with scale (CLIP 0.82$\rightarrow$0.87, SigLIP 0.74$\rightarrow$0.88, base$\rightarrow$XL), though even the strongest model stays below 0.90. We relate it to one measurable quantity, the median val$\rightarrow$A shift $\Delta$ in the APS true-class nonconformity score, which decreases with scale. The val threshold transfers only if target true-class scores are exchangeable with val's; on A they are systematically larger, consistent with the adversarial filtering of ImageNet-A. Across the tested checkpoints $\Delta$ is strongly negatively correlated with marginal coverage (Spearman $\rho$=-0.97 over the pooled R/A points, -1.00 over ImageNet-A alone; Fig.~\ref{fig:mechanism}). So scale partially recovers marginal coverage on A by shrinking $\Delta$ but never reaches target, while it never buys the Sketch conditional tail (main-paper Fig. 5). Full R/A ladder in Appendix~C.

\suppsection{Pretraining-corpus robustness}
Beyond scale, we vary the pretraining corpus, adding MetaCLIP [35], DataComp-1B [36], and DFN-2B [37] to CLIP/OpenCLIP/SigLIP: six ViT-B/16 VLMs spanning modern pretraining data, including MetaCLIP, the backbone Conf-OT and Fillioux et al.~[12] evaluate (full table Appendix~M). Every corpus reproduces the Sketch findings: APS-marginal coverage undercovers (0.843--0.866) and worst-class coverage collapses to 0.00 with 10--15\% of classes below the null floor for all six. The failure does not track ID accuracy or efficiency: DFN-2B is the most accurate (74.8\% ID) and most efficient (set size 7.4) checkpoint, yet has the largest below-null fraction (0.145). The ImageNet-A score-shift association (Sec.~S9) also holds across the 12 combined scale-and-corpus checkpoints ($\rho$=-0.99; Appendix~C). Within the tested ViT-B/16 sweep, neither pretraining corpus nor higher ID accuracy removes the class-tail failure.

\suppsection{Non-ImageNet replication}
A natural worry is that everything above lives in ImageNet's label space. We therefore run the core audit on two entirely different zero-shot benchmarks: Stanford Cars (196 fine-grained classes) [40] and Food-101 (101 classes) [41]: calibrating and evaluating in-distribution on each (10 splits, Appendix~N). The marginal-conditional gap reproduces the qualitative pattern the mechanism (Main Sec.~4.3) predicts: on Stanford Cars, CLIP attains 0.90 marginal coverage but only 0.18 worst-class coverage (vs a 0.56 null floor): the same gap, on a non-ImageNet domain, without any distribution shift, because fine-grained car recognition has heterogeneous per-class accuracy (CLIP 64\%). The effect is accuracy-gated, as Main Sec.~4.3 predicts: it is substantially milder for the higher-accuracy OpenCLIP and SigLIP (88--91\% on Cars) than for CLIP, with small below-null fractions (0.009 and 0.003), although their worst-class minima can still fall below the finite-sample floor (OpenCLIP 0.41, SigLIP 0.52, vs a 0.56 floor); on the easier Food-101 (88--92\%) all families sit near the floor with negligible below-null breadth. The family efficiency gap (Main Sec.~4.5) also reproduces (CLIP set size 3.3 vs SigLIP 1.2 on Cars). These results show that heterogeneous per-class difficulty can produce the marginal--tail gap outside ImageNet even without a source-to-target shift, in the tested settings where per-class accuracy is strongly heterogeneous.

\textbf{Under controlled corruption.} The deployment-shift collapse also reproduces off ImageNet. Applying a Gaussian-blur corruption to Cars ("Cars-C") under the controlled clean-to-blur covariate-shift protocol (calibrate APS-marginal on clean Cars, deploy on blurred Cars; Appendix~N) reproduces the ImageNet pattern qualitatively: accuracy drops sharply (CLIP 0.64$\rightarrow$0.31, OpenCLIP/SigLIP 0.88--0.91$\rightarrow$0.56), marginal coverage undercovers (0.79--0.86), the worst class falls to 0.17--0.34 (vs a 0.69 null floor), 10--25\% of classes drop below the floor, and set sizes explode (CLIP 24.8). Across blur severities 1$\rightarrow$5 the degradation is dose-dependent: the worst class stays collapsed (0.09--0.20) at every severity while marginal coverage holds ($\approx$0.91) until the harshest level, mirroring the ImageNet-C dose-response (Sec.~S5). Blurring the easy Food-101 (negligible below-null breadth in-distribution) similarly collapses 2 of 3 families (OpenCLIP undercovers to 0.81 with 52\% of classes below the floor, SigLIP 18\%) while CLIP's flatter scores over-cover, the same family-dependent pattern as Sec.~S9. These experiments identify two routes to class-tail failure: on a difficult fine-grained task, heterogeneous per-class accuracy can produce the gap even in-distribution, as for CLIP on Stanford Cars; on an easier task, corruption can induce the same gap, as on Food-C. Distribution shift is therefore one route to the failure; heterogeneous class difficulty is the broader mechanism.

\clearpage\section*{Appendices}\pdfbookmark[0]{Appendices}{appendices}
\pdfbookmark[1]{A. Conf-OT reproduction and hyperparameter sensitivity}{app.A}
\textbf{A. Conf-OT reproduction \& hyperparameter sensitivity.} Our clean-room reimplementation reproduces Conf-OT's in-distribution efficiency at valid coverage (APS set size -11 to -17\% across families, slightly below the paper's reported ~20\%). Sweeping the Sinkhorn temperature $\tau$ and iteration count on ImageNet-Sketch (CLIP, APS), marginal coverage and set size move substantially but the worst-class tail does not:

\begin{center}\small\setlength{\tabcolsep}{6pt}
\resizebox{\ifdim\width>\linewidth \linewidth\else\width\fi}{!}{%
\begin{tabular}{lrrrr}
\toprule
$\tau$ & n\_iter & marg cov & set size & worst-class cov \\
\midrule
(plain) & N/A & 0.862 & 26.7 & 0.000 \\
0.5 & 3 & 0.857 & 15.4 & 0.000 \\
1.0 & 3 & 0.896 & 27.6 & 0.000 \\
1.0 & 10 & 0.902 & 28.3 & 0.020 \\
2.0 & 5 & 0.907 & 33.0 & 0.020 \\
\bottomrule
\end{tabular}}
\end{center}
Verification against the authors' code: running the authors' released Sinkhorn routine (jusiro/CLIP-Conformal) on our cached CLIP logits also yields worst-class coverage 0.00 on Sketch, so the tail failure is not an artifact of our reimplementation or tested settings. Efficiency is not directly comparable (we feed cosine-scaled similarities to avoid overflow rather than replicating their full pipeline), so the -11--17\% figures are our reimplementation's reference only.

\pdfbookmark[1]{B. alpha-robustness}{app.B}
\textbf{B. $\alpha$-robustness.} Repeating the audit at $\alpha$=0.05 (target 0.95) on Sketch: the collapse and marginal undercoverage persist (full table):

\begin{center}\small\setlength{\tabcolsep}{6pt}
\resizebox{\ifdim\width>\linewidth \linewidth\else\width\fi}{!}{%
\begin{tabular}{lrrrrr}
\toprule
family & marg cov & set size & null floor & frac<null & p5 \\
\midrule
CLIP & 0.928 & 49.1 & 0.78 & 0.074 & 0.73 \\
OpenCLIP & 0.925 & 25.3 & 0.78 & 0.092 & 0.71 \\
SigLIP & 0.926 & 19.2 & 0.78 & 0.069 & 0.73 \\
\bottomrule
\end{tabular}}
\end{center}
Worst-class $\alpha$-frontier (CLIP, Sketch). Tightening the target does not fix the worst class: it only inflates sets:

\begin{center}\small\setlength{\tabcolsep}{6pt}
\resizebox{\ifdim\width>\linewidth \linewidth\else\width\fi}{!}{%
\begin{tabular}{lrrr}
\toprule
target 1-$\alpha$ & marg cov & worst-class & mean set size \\
\midrule
0.900 & 0.862 & 0.00 & 26.7 \\
0.990 & 0.984 & 0.02 & 139.8 \\
0.995 & 0.992 & 0.08 & 209.7 \\
0.999 & 0.998 & 0.31 & 453.4 \\
\bottomrule
\end{tabular}}
\end{center}
Even at target 0.999 the worst class reaches only 0.31, with sets of 453/1000 classes, so worst-class safety via marginal tightening is prohibitively expensive; the same holds for all three families.

\pdfbookmark[1]{C. Scale ladder (full)}{app.C}
\textbf{C. Scale ladder (full).} Base$\rightarrow$large$\rightarrow$XL $\times$ 3 families on ImageNet-V2 and -Sketch (full table). On Sketch, accuracy rises sharply with scale while marginal coverage and below-null fraction stay flat:

\begin{center}\small\setlength{\tabcolsep}{6pt}
\resizebox{\ifdim\width>\linewidth \linewidth\else\width\fi}{!}{%
\begin{tabular}{lrrrr}
\toprule
family & base$\rightarrow$large$\rightarrow$XL & Sketch acc & marg cov & frac<null \\
\midrule
CLIP & B/16$\rightarrow$L/14$\rightarrow$L-336 & 0.48$\rightarrow$0.60$\rightarrow$0.61 & 0.86$\rightarrow$0.85$\rightarrow$0.85 & 0.12$\rightarrow$0.12$\rightarrow$0.11 \\
OpenCLIP & B/16$\rightarrow$L/14$\rightarrow$H/14 & 0.56$\rightarrow$0.63$\rightarrow$0.67 & 0.86$\rightarrow$0.86$\rightarrow$0.86 & 0.12$\rightarrow$0.11$\rightarrow$0.13 \\
SigLIP & base$\rightarrow$large$\rightarrow$so400m & 0.66$\rightarrow$0.71$\rightarrow$0.73 & 0.87$\rightarrow$0.87$\rightarrow$0.86 & 0.10$\rightarrow$0.10$\rightarrow$0.11 \\
\bottomrule
\end{tabular}}
\end{center}
ImageNet-R/-A ladder (restricted-200, APS, $\alpha$=0.10). On R, coverage is essentially valid at every scale; on A, the marginal coverage collapses and rises with scale but never reaches target: the opposite of the scale-persistent conditional tail on Sketch:

\begin{center}\small\setlength{\tabcolsep}{6pt}
\resizebox{\ifdim\width>\linewidth \linewidth\else\width\fi}{!}{%
\begin{tabular}{lrrrr}
\toprule
family & base$\rightarrow$large$\rightarrow$XL & R marg cov & A acc & A marg cov \\
\midrule
CLIP & B/16$\rightarrow$L/14$\rightarrow$L-336 & 0.90$\rightarrow$0.90$\rightarrow$0.91 & 0.50$\rightarrow$0.71$\rightarrow$0.77 & 0.82$\rightarrow$0.86$\rightarrow$0.87 \\
OpenCLIP & B/16$\rightarrow$L/14$\rightarrow$H/14 & 0.88$\rightarrow$0.89$\rightarrow$0.89 & 0.38$\rightarrow$0.54$\rightarrow$0.59 & 0.65$\rightarrow$0.73$\rightarrow$0.74 \\
SigLIP & base$\rightarrow$large$\rightarrow$so400m & 0.91$\rightarrow$0.91$\rightarrow$0.91 & 0.44$\rightarrow$0.61$\rightarrow$0.80 & 0.74$\rightarrow$0.82$\rightarrow$0.88 \\
\bottomrule
\end{tabular}}
\end{center}
Mechanism of the A-collapse. The median val$\rightarrow$A shift $\Delta$ in the APS true-class nonconformity score is strongly negatively correlated with marginal coverage (Spearman $\rho$=-1.00 across the 9 A points; -0.97 pooled over R+A) and shrinks monotonically with scale (on R, $\Delta$$\approx$0 and coverage holds):

\begin{center}\small\setlength{\tabcolsep}{6pt}
\resizebox{\ifdim\width>\linewidth \linewidth\else\width\fi}{!}{%
\begin{tabular}{lrrr}
\toprule
family & base$\rightarrow$large$\rightarrow$XL & A: $\Delta$ score & A: marg cov \\
\midrule
CLIP & B/16$\rightarrow$L/14$\rightarrow$L-336 & +0.087$\rightarrow$+0.044$\rightarrow$+0.020 & 0.82$\rightarrow$0.86$\rightarrow$0.87 \\
OpenCLIP & B/16$\rightarrow$L/14$\rightarrow$H/14 & +0.272$\rightarrow$+0.178$\rightarrow$+0.160 & 0.65$\rightarrow$0.73$\rightarrow$0.74 \\
SigLIP & base$\rightarrow$large$\rightarrow$so400m & +0.172$\rightarrow$+0.087$\rightarrow$+0.014 & 0.74$\rightarrow$0.82$\rightarrow$0.88 \\
\bottomrule
\end{tabular}}
\end{center}
\pdfbookmark[1]{D. Protocol details}{app.D}
\textbf{D. Protocol details.} Logits are extracted once per (model, dataset) and cached, so all conformal analyses run from fixed logits; engine correctness is checked by exact ID coverage, analytic Vovk-band tests, and synthetic-logit unit tests (Sec.~3). The logit inventory covers 9 checkpoints (3 families $\times$ base/large/XL) over {val, V2, Sketch, R, A}, ImageNet-C over 3 corruption types $\times$ 5 severities, and a matched single-prompt cache (3 base families $\times$ {val, Sketch}). Protocol: $\alpha$=0.10 primary; 10 calibration/test splits with bootstrap 95\% confidence intervals; the null-MCCC band is estimated by simulation under the observed per-class counts; ImageNet-R/-A use the restricted-200 protocol; SigLIP is softmax-normalized for cross-family comparison.

\textbf{Null-band simulation (details).} The simulated null-MCCC band assumes every class is covered i.i.d. at the nominal rate $1{-}\alpha{=}0.90$: for each class $y$ with observed test count $n_y{\ge}5$ we draw an independent $\mathrm{Binomial}(n_y, 0.90)/n_y$, take the minimum over classes, and repeat for $n_{\text{sim}}{=}10000$ Monte-Carlo trials under a fixed seed. The reported null floor is the 2.5th percentile of this minimum-coverage distribution (e.g. $\approx$0.70 at 50 img/class, $\approx$0.79 at 150/class), and a class is counted below-null if its observed coverage falls under that floor. Under the reference null, only 2.5\% of complete $K$-class simulations have a minimum below this floor; the fraction of observed classes below the same threshold is a descriptive measure of the breadth of tail failure, not itself a calibrated per-class error probability. Classes with $n_y{<}5$ are excluded from MCCC (too few examples to estimate per-class coverage), which is why we lead with the dense Sketch/R sets and report the percentile (p5/p10/below-null) statistics rather than the raw worst-of-$K$ minimum.

\pdfbookmark[1]{E. Prompt-template ablation (the cross-family set-size gap is not a prompt artifact)}{app.E}
\textbf{E. Prompt-template ablation (the cross-family set-size gap is not a prompt artifact).} Re-extracting all three base families with one matched single prompt (\texttt{"a photo of a \{\}."}), replacing CLIP/OpenCLIP's 80-template ensemble and SigLIP's canonical prompt, on ImageNet-val + -Sketch. The 80-template ensemble sharpens CLIP/OpenCLIP and shrinks their sets, so removing it inflates them while SigLIP (already single-prompt) is unchanged: the family ordering persists and the efficiency gap widens:

\begin{center}\small\setlength{\tabcolsep}{6pt}
\resizebox{\ifdim\width>\linewidth \linewidth\else\width\fi}{!}{%
\begin{tabular}{lrrr}
\toprule
family & ID size (ens$\rightarrow$single) & Sketch size (ens$\rightarrow$single) & Sketch acc (ens$\rightarrow$single) \\
\midrule
CLIP & 11.6 $\rightarrow$ 13.8 & 26.8 $\rightarrow$ 29.2 & 0.483 $\rightarrow$ 0.461 \\
OpenCLIP & 5.6 $\rightarrow$ 6.1 & 14.3 $\rightarrow$ 15.4 & 0.561 $\rightarrow$ 0.548 \\
SigLIP & 5.1 $\rightarrow$ 4.9 & 9.4 $\rightarrow$ 9.1 & 0.663 $\rightarrow$ 0.666 \\
\bottomrule
\end{tabular}}
\end{center}
Sketch set-size ratio vs the most-efficient family: ensemble CLIP 2.84$\times$ / OpenCLIP 1.51$\times$; matched single prompt CLIP 3.21$\times$ / OpenCLIP 1.70$\times$, so the differing prompt protocol, if anything, understated CLIP's inefficiency. Marginal undercoverage (0.858--0.865) and below-null fraction (0.108--0.130) on Sketch are also unchanged under the matched prompt, so neither the efficiency gap nor the conditional collapse is a prompt artifact.

\pdfbookmark[1]{F. Cross-shift calibration transfer (conservative harder-proxy)}{app.F}
\textbf{F. Cross-shift calibration transfer (conservative harder-proxy).} APS-marginal coverage (target 0.90) when the conformal threshold is calibrated on the row shift and deployed on the column shift, over the shared 1000-class space {val, V2, Sketch}. Diagonal = same-shift split ($\approx$ oracle). Calibrating on a shift at least as severe as the target over-covers (lower-left); a milder shift under-covers (upper-right). The structure is identical across families: a harder proxy shift is a deployable remedy for marginal coverage (not the conditional tail). ImageNet-R/-A are omitted (their 200-class subsets share only 86/200 classes).

\begin{center}\small\setlength{\tabcolsep}{6pt}
\resizebox{\ifdim\width>\linewidth \linewidth\else\width\fi}{!}{%
\begin{tabular}{lrrr}
\toprule
calibrate $\downarrow$ / deploy $\rightarrow$ & val & V2 & Sketch \\
\midrule
CLIP: val & 0.899 & 0.887 & 0.859 \\
CLIP: V2 & 0.910 & 0.900 & 0.878 \\
CLIP: Sketch & 0.929 & 0.919 & 0.901 \\
OpenCLIP: val & 0.904 & 0.870 & 0.858 \\
OpenCLIP: V2 & 0.926 & 0.893 & 0.892 \\
OpenCLIP: Sketch & 0.932 & 0.906 & 0.897 \\
SigLIP: val & 0.900 & 0.878 & 0.862 \\
SigLIP: V2 & 0.919 & 0.893 & 0.886 \\
SigLIP: Sketch & 0.931 & 0.911 & 0.900 \\
\bottomrule
\end{tabular}}
\end{center}
\pdfbookmark[1]{G. RAPS sensitivity: kreg has limited effect, lambda controls the coverage--efficiency trade-off}{app.G}
\textbf{G. RAPS sensitivity: $k_{\text{reg}}$ has limited effect, $\lambda$ controls the coverage--efficiency trade-off.} Marginal coverage / mean set size on ImageNet-Sketch (APS baseline for reference). RAPS shrinks the APS set at every $(k_{\text{reg}},\lambda)$; the trade-off is $\lambda$-governed ($k_{\text{reg}}$ barely matters), and the native-SigLIP set-size reduction holds for all $\lambda\ge0.01$. The recommendation should not be read as tuning-robust to $\lambda$, which controls the coverage--efficiency trade-off.

\begin{center}\small\setlength{\tabcolsep}{6pt}
\resizebox{\ifdim\width>\linewidth \linewidth\else\width\fi}{!}{%
\begin{tabular}{lrrrr}
\toprule
family (APS baseline) & $\lambda{=}0.001$ & $\lambda{=}0.01$ & $\lambda{=}0.05$ & $\lambda{=}0.1$ \\
\midrule
CLIP (0.862 / 26.7) & 0.849 / 17.2 & 0.813 / 8.4 & 0.770 / 4.9 & 0.758 / 4.4 \\
OpenCLIP (0.862 / 14.3) & 0.853 / 9.0 & 0.830 / 5.1 & 0.811 / 4.1 & 0.806 / 4.0 \\
SigLIP-softmax (0.866 / 9.4) & 0.863 / 6.8 & 0.851 / 4.0 & 0.836 / 2.7 & 0.830 / 2.6 \\
SigLIP-sigmoid (0.988 / 940.5) & 0.986 / 618.9 & 0.960 / 64.4 & 0.930 / 14.5 & 0.907 / 8.3 \\
\bottomrule
\end{tabular}}
\end{center}
(All at $k_{\text{reg}}{=}1$; the full $k_{\text{reg}}\in\{1,2,5,10\}$ grid moves set size by at most 2 and coverage by at most 0.03.) The sigmoid row shows why $\lambda\ge0.01$ matters: at $\lambda{=}0.001$ RAPS barely regularizes the ill-posed sigmoid-APS sets (619), but $\lambda{=}0.05$ tames them to 14.5: versus APS's 940.

\pdfbookmark[1]{H. Size-stratified coverage (large sets are less informative and less covered)}{app.H}
\textbf{H. Size-stratified coverage (large sets are less informative and less covered).} APS-marginal coverage on ImageNet-Sketch within bins of prediction-set size. Coverage is highest in the small-set bins and lowest in the largest-set regime, although intermediate bins are not strictly monotone; OpenCLIP and SigLIP singletons exceed the 0.90 target, while CLIP singletons are near nominal at 0.894. The trend is sharpest for SigLIP and mildest for CLIP, mirroring their set-size distributions (Main Sec.~4.5).

\begin{center}\small\setlength{\tabcolsep}{6pt}
\resizebox{\ifdim\width>\linewidth \linewidth\else\width\fi}{!}{%
\begin{tabular}{lrrrrr}
\toprule
family (overall cov / mean size) & 1 (singleton) & 2--5 & 6--20 & 21--100 & 101+ \\
\midrule
CLIP (0.862 / 26.7) & 0.894 (18\%) & 0.872 (29\%) & 0.856 (27\%) & 0.870 (19\%) & 0.848 (7\%) \\
OpenCLIP (0.862 / 14.3) & 0.911 (29\%) & 0.884 (34\%) & 0.837 (25\%) & 0.782 (10\%) & 0.817 (2\%) \\
SigLIP (0.866 / 9.4) & 0.919 (41\%) & 0.884 (33\%) & 0.846 (15\%) & 0.816 (8\%) & 0.589 (2\%) \\
\bottomrule
\end{tabular}}
\end{center}
(Parenthetical = fraction of test examples in that bin.) Singletons sit above the 0.90 target for every family; the largest sets fall furthest below, so the marginal number is an average over an over-covered head and an under-covered tail, and set size is an inverted reliability signal under shift.

\pdfbookmark[1]{I. Size-thresholded selective conformal prediction (deployable recipe)}{app.I}
\textbf{I. Size-thresholded selective conformal prediction (deployable recipe).} Defer when the APS set size exceeds cap $k$; act on the retained (small-set) examples. ImageNet-Sketch, calibrated marginally on clean val. Retention = fraction acted on; the reported coverage and top-1 are empirical retained values, computed over the retained non-empty prediction sets (selection changes the evaluated population, so these are not finite-sample conformal guarantees).

\begin{center}\small\setlength{\tabcolsep}{6pt}
\resizebox{\ifdim\width>\linewidth \linewidth\else\width\fi}{!}{%
\begin{tabular}{lrrrrr}
\toprule
 & $k{\le}1$ & $k{\le}3$ & $k{\le}5$ & $k{\le}10$ & all \\
\midrule
CLIP retain / cov / acc & 18\% / 0.89 / 0.89 & 37\% / 0.89 / 0.75 & 47\% / 0.88 / 0.69 & 62\% / 0.87 / 0.62 & 99\% / 0.87 / 0.48 \\
OpenCLIP retain / cov / acc & 29\% / 0.91 / 0.91 & 51\% / 0.90 / 0.78 & 62\% / 0.90 / 0.73 & 76\% / 0.89 / 0.66 & 99\% / 0.87 / 0.56 \\
SigLIP retain / cov / acc & 41\% / 0.92 / 0.92 & 65\% / 0.91 / 0.81 & 74\% / 0.90 / 0.77 & 83\% / 0.90 / 0.73 & 98\% / 0.88 / 0.66 \\
\bottomrule
\end{tabular}}
\end{center}
Lowering $k$ raises both coverage and accuracy on the retained set, monotonically; SigLIP dominates (most retention at any coverage). Deferring the ~26\% hardest inputs ($k{\le}5$) restores SigLIP to nominal 0.90 coverage at 0.77 top-1 (vs 0.66 over all inputs).

Conf-OT $\times$ selective compose. Applying the size threshold to Conf-OT's transported codes instead of plain softmax raises retained coverage $\approx$2--3 pts at the same retention/set size: the strongest label-free recipe:

\begin{center}\small\setlength{\tabcolsep}{6pt}
\resizebox{\ifdim\width>\linewidth \linewidth\else\width\fi}{!}{%
\begin{tabular}{lrr}
\toprule
family & plain+sel (k$\leq$5) ret/cov & Conf-OT+sel (k$\leq$5) ret/cov \\
\midrule
CLIP & 47\% / 0.880 & 45\% / 0.912 \\
OpenCLIP & 62\% / 0.896 & 61\% / 0.919 \\
SigLIP & 74\% / 0.904 & 74\% / 0.922 \\
\bottomrule
\end{tabular}}
\end{center}
\pdfbookmark[1]{J. Methods comparison, all three families (the body's decision table, full)}{app.J}
\textbf{J. Methods comparison, all three families (the body's decision table, full).} Every method on ImageNet-Sketch, one shared calibration split per family; CLIP is the version shown in the body.

\begin{center}\small\setlength{\tabcolsep}{6pt}
\resizebox{\ifdim\width>\linewidth \linewidth\else\width\fi}{!}{%
\begin{tabular}{lrrrrrr}
\toprule
family & method & target info & retention & marg cov & worst-class & set size \\
\midrule
CLIP & plain & none & 100\% & 0.86 & 0.00 & 26.7 \\
CLIP & source-Mondrian & none & 100\% & 0.87 & 0.00 & 45.3 \\
CLIP & clustered & none & 100\% & 0.83 & 0.00 & 29.6 \\
CLIP & Conf-OT & unlabeled & 100\% & 0.90 & 0.00 & 27.9 \\
CLIP & selective (k$\leq$5) & none & 47\% & 0.88 & N/A & 2.3 \\
CLIP & oracle & labeled & 100\% & 0.91 & 0.75 & 39.2 \\
OpenCLIP & plain & none & 100\% & 0.86 & 0.00 & 14.3 \\
OpenCLIP & source-Mondrian & none & 100\% & 0.87 & 0.00 & 29.9 \\
OpenCLIP & clustered & none & 100\% & 0.83 & 0.00 & 17.7 \\
OpenCLIP & Conf-OT & unlabeled & 100\% & 0.89 & 0.00 & 14.7 \\
OpenCLIP & selective (k$\leq$5) & none & 62\% & 0.90 & N/A & 2.1 \\
OpenCLIP & oracle & labeled & 100\% & 0.91 & 0.76 & 23.7 \\
SigLIP & plain & none & 100\% & 0.87 & 0.00 & 9.4 \\
SigLIP & source-Mondrian & none & 100\% & 0.88 & 0.00 & 20.3 \\
SigLIP & clustered & none & 100\% & 0.84 & 0.00 & 11.6 \\
SigLIP & Conf-OT & unlabeled & 100\% & 0.89 & 0.00 & 9.2 \\
SigLIP & selective (k$\leq$5) & none & 74\% & 0.90 & N/A & 1.8 \\
SigLIP & oracle & labeled & 100\% & 0.91 & 0.75 & 17.8 \\
\bottomrule
\end{tabular}}
\end{center}
The pattern is family-invariant: only the oracle substantially lifts the worst-class tail (to a sub-nominal $\approx$0.75); Conf-OT restores marginal coverage (unlabeled target); selective prediction empirically recovers near-nominal retained coverage label-free on a retained 47--74\% of inputs. The oracle's $\approx$0.75 is a finite-sample sub-nominal value at the available target-label budget, not a full restoration to 0.90 (Sec.~S4 traces the budget curve).

\pdfbookmark[1]{K. Repair-baseline configurations (fair-comparison summary)}{app.K}
\textbf{K. Repair-baseline configurations (fair-comparison summary).} To preempt questions about whether baselines are configured fairly, we summarize each method's information access and key settings (all on the same shared calibration split per family):

\begin{center}\small\setlength{\tabcolsep}{6pt}
\resizebox{\ifdim\width>\linewidth \linewidth\else\width\fi}{!}{%
\begin{tabular}{lrrrr}
\toprule
method & target labels? & uses unlabeled target? & key hyperparameters & purpose \\
\midrule
plain split & no & no & $\alpha$ & baseline marginal CP \\
source-Mondrian & no & no & per-class source quantile & source-side class-conditional CP \\
clustered & no & no & \#clusters, class-similarity & sparse-class pooling [6] \\
Conf-OT & no & yes & $\tau$, iterations, $\varepsilon$ (Sinkhorn) & transductive marginal/average repair \\
pseudo-Mondrian & no & yes (pseudo-labels) & argmax pseudo-label & label-free per-class attempt \\
selective & no & no & set-size threshold $k$ & empirical retained-coverage abstention \\
oracle & yes & no & target per-class quantile & non-deployable upper bound \\
\bottomrule
\end{tabular}}
\end{center}
All no-target-label methods leave the worst-class tail at or near zero; only the labeled-target oracle lifts it (and only to a sub-nominal $\approx$0.75). Conf-OT is swept over its Sinkhorn settings (Appendix~A) and verified against the authors' released code; clustered conformal over its cluster count; the tail result is stable in every configuration.

\pdfbookmark[1]{L. Abstention signal: conformal set size vs. confidence}{app.L}
\textbf{L. Abstention signal: conformal set size vs. confidence.} At matched retention on ImageNet-Sketch, deferring the largest-set vs. the least-confident examples gives near-identical conformal coverage (within $\approx$0.01): set size and top-1 confidence are substitutable abstention signals (negatively correlated -0.36 to -0.56); confidence holds a slight edge on retained accuracy. We use set size because it is the native conformal output and directly controls the retained prediction-set complexity, not because it ranks better.

\pdfbookmark[1]{M. Six pretraining corpora (corpus-invariance)}{app.M}
\textbf{M. Six pretraining corpora (corpus-invariance).} APS audit on ImageNet-Sketch across six ViT-B/16 VLMs spanning the diversity of modern pretraining data; MetaCLIP is the backbone Conf-OT/Fillioux'24 use. All undercover marginally and collapse to worst-class 0.00 with 10--15\% of classes below the null floor: independent of corpus, accuracy, or efficiency.

\begin{center}\small\setlength{\tabcolsep}{6pt}
\resizebox{\ifdim\width>\linewidth \linewidth\else\width\fi}{!}{%
\begin{tabular}{lrrrrrrr}
\toprule
family & corpus & ID acc & Sketch acc & marg cov & set size & worst-class & frac<null \\
\midrule
CLIP & OpenAI-WIT 400M & 0.675 & 0.483 & 0.862 & 26.7 & 0.00 & 0.118 \\
OpenCLIP & LAION-2B & 0.695 & 0.561 & 0.862 & 14.3 & 0.00 & 0.122 \\
SigLIP & WebLI & 0.754 & 0.663 & 0.866 & 9.4 & 0.00 & 0.102 \\
MetaCLIP & CommonCrawl-2.5B & 0.727 & 0.602 & 0.862 & 16.3 & 0.00 & 0.108 \\
DataComp & DataComp-1B & 0.730 & 0.605 & 0.854 & 10.1 & 0.00 & 0.126 \\
DFN & DFN-2B & 0.748 & 0.614 & 0.843 & 7.4 & 0.00 & 0.145 \\
\bottomrule
\end{tabular}}
\end{center}
Split-stability. Over 10 random calibration/test splits, every family's marginal-coverage CI lies entirely below 0.90 and its below-null-fraction CI excludes 0 (e.g. DFN 0.843 [0.840,0.846] cov, 0.145 [0.139,0.150] below-null; worst-class 0.000 throughout): the across-corpus collapse is statistically stable, not a single-seed artifact.

ImageNet-A marginal coverage + val$\rightarrow$A score shift $\Delta$ (all six show the same score-shift association; pooled with the scale ladder, $\Delta$ is negatively correlated with marginal coverage at $\rho$=-0.99 over 12 checkpoints):

\begin{center}\small\setlength{\tabcolsep}{6pt}
\resizebox{\ifdim\width>\linewidth \linewidth\else\width\fi}{!}{%
\begin{tabular}{lrrr}
\toprule
family & A acc & marg cov & $\Delta$ score \\
\midrule
CLIP & 0.501 & 0.820 & +0.087 \\
OpenCLIP & 0.383 & 0.645 & +0.272 \\
SigLIP & 0.440 & 0.741 & +0.172 \\
MetaCLIP & 0.496 & 0.828 & +0.084 \\
DataComp & 0.484 & 0.751 & +0.161 \\
DFN & 0.468 & 0.696 & +0.212 \\
\bottomrule
\end{tabular}}
\end{center}
\pdfbookmark[1]{N. Out-of-ImageNet replication (Stanford Cars, Food-101)}{app.N}
\textbf{N. Out-of-ImageNet replication (Stanford Cars, Food-101).} Core audit on two non-ImageNet zero-shot benchmarks, calibrated + evaluated in-distribution (10 splits). The marginal-conditional gap appears on the hard fine-grained task (Cars, where per-class accuracy is heterogeneous) and is accuracy-gated (absent on easy Food-101) confirming the Main Sec.~4.3 mechanism off ImageNet. Family efficiency gap (Main Sec.~4.5) also reproduces (CLIP set size > SigLIP on Cars).

\begin{center}\small\setlength{\tabcolsep}{6pt}
\resizebox{\ifdim\width>\linewidth \linewidth\else\width\fi}{!}{%
\begin{tabular}{lrrrrrrr}
\toprule
dataset & family & acc & marg cov & set size & worst-class & null floor & frac<null \\
\midrule
Stanford Cars (196) & CLIP & 0.644 & 0.902 & 3.3 & 0.18 & 0.56 & 0.030 \\
Stanford Cars (196) & OpenCLIP & 0.881 & 0.904 & 1.4 & 0.41 & 0.56 & 0.009 \\
Stanford Cars (196) & SigLIP & 0.908 & 0.904 & 1.2 & 0.52 & 0.56 & 0.003 \\
Food-101 (101) & CLIP & 0.884 & 0.903 & 1.6 & 0.80 & 0.80 & 0.004 \\
Food-101 (101) & OpenCLIP & 0.865 & 0.902 & 1.5 & 0.78 & 0.80 & 0.009 \\
Food-101 (101) & SigLIP & 0.915 & 0.901 & 1.4 & 0.81 & 0.80 & 0.002 \\
\bottomrule
\end{tabular}}
\end{center}
On Cars, worst-class coverage tracks accuracy (CLIP 0.64 acc $\rightarrow$ 0.18; SigLIP 0.91 acc $\rightarrow$ 0.52 $\approx$ floor); on the easy Food-101 all families have negligible below-null breadth: the gap is accuracy-gated, not label-space-specific.

Cars-C deployment shift. Calibrate APS-marginal on clean Cars, deploy on Gaussian-blurred Cars (severity 5): a controlled blur covariate shift on a non-ImageNet label space (10 splits). The full ImageNet pattern reproduces: marginal undercoverage + worst-class collapse below the null floor.

\begin{center}\small\setlength{\tabcolsep}{6pt}
\resizebox{\ifdim\width>\linewidth \linewidth\else\width\fi}{!}{%
\begin{tabular}{lrrrrrr}
\toprule
family & clean$\rightarrow$blur acc & marg cov & set size & worst-class & null floor & frac<null \\
\midrule
CLIP & 0.64$\rightarrow$0.31 & 0.857 & 24.8 & 0.17 & 0.69 & 0.097 \\
OpenCLIP & 0.88$\rightarrow$0.57 & 0.792 & 6.7 & 0.31 & 0.69 & 0.199 \\
SigLIP & 0.91$\rightarrow$0.56 & 0.785 & 10.7 & 0.34 & 0.69 & 0.247 \\
\bottomrule
\end{tabular}}
\end{center}
Cars-C severity curve (CLIP). Calibrate clean Cars $\rightarrow$ deploy blurred at severity 1--5; the collapse is dose-dependent (set size explodes; worst-class collapsed at every level; marginal holds until the harshest):

\begin{center}\small\setlength{\tabcolsep}{6pt}
\resizebox{\ifdim\width>\linewidth \linewidth\else\width\fi}{!}{%
\begin{tabular}{lrrrrr}
\toprule
blur severity & acc & marg cov & set size & worst-class & frac<null \\
\midrule
1 & 0.622 & 0.908 & 3.8 & 0.20 & 0.051 \\
2 & 0.564 & 0.912 & 5.7 & 0.19 & 0.052 \\
3 & 0.494 & 0.906 & 8.8 & 0.09 & 0.047 \\
4 & 0.424 & 0.897 & 13.2 & 0.12 & 0.042 \\
5 & 0.305 & 0.857 & 24.8 & 0.17 & 0.097 \\
\bottomrule
\end{tabular}}
\end{center}
(Null floor 0.69 throughout.) Set size and accuracy degrade monotonically with severity; worst-class coverage is collapsed at all severities.

Food-C : an easy dataset collapses once shifted. Calibrate clean Food-101 $\rightarrow$ deploy Gaussian-blurred. Food-101 has negligible in-distribution below-null breadth (Sec.~S11), but under shift 2/3 families collapse (null floor 0.83):

\begin{center}\small\setlength{\tabcolsep}{6pt}
\resizebox{\ifdim\width>\linewidth \linewidth\else\width\fi}{!}{%
\begin{tabular}{lrrrrr}
\toprule
family & clean$\rightarrow$blur acc & marg cov & set size & worst-class & frac<null \\
\midrule
CLIP & 0.88$\rightarrow$0.59 & 0.931 & 11.8 & 0.81 & 0.010 \\
OpenCLIP & 0.87$\rightarrow$0.61 & 0.807 & 3.7 & 0.51 & 0.518 \\
SigLIP & 0.92$\rightarrow$0.60 & 0.881 & 8.5 & 0.72 & 0.179 \\
\bottomrule
\end{tabular}}
\end{center}
CLIP over-covers (flat scores, negligible below-null breadth); OpenCLIP/SigLIP undercover and collapse (52\%/18\% below null): the family-dependent marginal pattern of Main Sec.~4.1/Sec.~S9, confirming the corruption-induced collapse is not Cars-specific.

\pdfbookmark[1]{O. Label-budget on a milder shift (ImageNet-R), confirming the cost is general (Sec.~S4)}{app.O}
\textbf{O. Label-budget on a milder shift (ImageNet-R), confirming the cost is general (Sec.~S4).} The Sec.~S4 target-label-budget sweep, repeated on ImageNet-R (restricted-200, $\approx$150 img/class: a milder class-conditional collapse than Sketch), 5 seeds. CLIP shown; OpenCLIP/SigLIP are near-identical (worst-class within $\approx$0.03, below-null within $\approx$0.02 at every budget).

\begin{center}\small\setlength{\tabcolsep}{6pt}
\resizebox{\ifdim\width>\linewidth \linewidth\else\width\fi}{!}{%
\begin{tabular}{lrrr}
\toprule
labels/class & worst-class & frac<null & marg cov \\
\midrule
1 & 1.00 $\pm$ 0.00 & 0.000 & 1.000 \\
3 & 1.00 $\pm$ 0.00 & 0.000 & 1.000 \\
5 & 1.00 $\pm$ 0.00 & 0.000 & 1.000 \\
10 & 0.55 $\pm$ 0.05 & 0.064 & 0.905 \\
20 & 0.61 $\pm$ 0.05 & 0.026 & 0.905 \\
30 & 0.67 $\pm$ 0.03 & 0.007 & 0.900 \\
\bottomrule
\end{tabular}}
\end{center}
The curve is qualitatively identical to Sketch (Sec.~S4): at $n{\le}5$ labels/class the conformal rank exceeds the per-class sample, so the threshold is infinite and coverage is trivially 1.0 with full sets; finite-threshold worst-class recovery then proceeds gradually and sub-nominally, needing $\approx$20--30 labels/class to drive the below-null fraction toward 0. A clean severity-vs-cost isolation is confounded (R has 3$\times$ the per-class samples and 1/5 the label space of Sketch, both of which shift the null band and the worst-of-$K$ minimum independent of severity), so we read this as generality: the prohibitive per-class label cost is a sample-complexity property of target-recalibration under shift rather than a severity law.

\pdfbookmark[1]{P. Open-vocabulary vocabulary-expansion: full dose-response (Sec.~S8)}{app.P}
\textbf{P. Open-vocabulary vocabulary-expansion: full dose-response (Sec.~S8).} Seen-class marginal coverage ($\alpha$=0.10) as the deployed vocabulary expands from $k_{\text{seen}}$ to the full 1000 classes (novel text-defined distractors added; truth always a seen class), softmax scores, mean over 10 partitions. control = no expansion ($\approx$0.90 throughout); oracle = recalibrated under the expanded vocabulary (restores $\approx$0.90). Penalty = control - expanded.

\begin{center}\small\setlength{\tabcolsep}{6pt}
\resizebox{\ifdim\width>\linewidth \linewidth\else\width\fi}{!}{%
\begin{tabular}{lrrrr}
\toprule
family & score & added & expanded cov & penalty \\
\midrule
CLIP & LAC & 250 / 500 / 750 & 0.881 / 0.852 / 0.777 & 0.018 / 0.050 / 0.124 \\
CLIP & APS & 250 / 500 / 750 & 0.899 / 0.899 / 0.900 & 0.000 / 0.001 / 0.002 \\
OpenCLIP & LAC & 250 / 500 / 750 & 0.884 / 0.858 / 0.795 & 0.015 / 0.043 / 0.106 \\
OpenCLIP & APS & 250 / 500 / 750 & 0.893 / 0.882 / 0.864 & 0.007 / 0.019 / 0.037 \\
SigLIP & LAC & 250 / 500 / 750 & 0.882 / 0.857 / 0.794 & 0.017 / 0.045 / 0.107 \\
SigLIP & APS & 250 / 500 / 750 & 0.894 / 0.892 / 0.885 & 0.005 / 0.010 / 0.016 \\
\bottomrule
\end{tabular}}
\end{center}
LAC's seen-class coverage falls in a clean dose-dependent way (penalty 0.106--0.124 at +750 across families) as novel distractors dilute its absolute-probability threshold via softmax renormalization; APS is substantially more stable (maximum penalty 0.037, at +750), because its cumulative-mass-above-truth score is largely unchanged when distractors rank below the true class. The score-dependence (not the expansion per se) is the finding (Sec.~S8).

\pdfbookmark[1]{Q. Positioning relative to prior work}{app.Q}
\textbf{Q. Positioning relative to prior work.} This table positions the audit relative to the closest conformal-VLM and shift-conformal work ($\checkmark$ = covered; partial / $\times$ as marked).

\begin{center}\small\setlength{\tabcolsep}{6pt}
\resizebox{\ifdim\width>\linewidth \linewidth\else\width\fi}{!}{%
\begin{tabular}{lrrrrr}
\toprule
work & VLM family span & SigLIP score geometry & shift axes & worst-tail audit & repair-tail analysis \\
\midrule
Fillioux et al. '24 & partial (CLIP/MetaCLIP) & $\times$ & natural & partial & $\times$ \\
Conf-OT '25 & CLIP-style & $\times$ & natural & average & partial \\
WQLCP '25 & $\times$ (trained nets) & $\times$ & natural & limited & method paper \\
LATA '26 & medical VLMs & $\times$ & domain & $\checkmark$ & $\checkmark$ (medical) \\
This work & CLIP/OpenCLIP/SigLIP & $\checkmark$ & natural+C+vocab & worst-tail + null band & $\checkmark$ \\
\bottomrule
\end{tabular}}
\end{center}

\begin{thebibliography}{99}
\bibitem{ref1}\hypertarget{bib.ref1}{} Vovk, Gammerman, Shafer. \emph{Algorithmic Learning in a Random World.} 2005.
\bibitem{ref2}\hypertarget{bib.ref2}{} Angelopoulos, Bates. \emph{A Gentle Introduction to Conformal Prediction and Distribution-Free Uncertainty Quantification.} arXiv:2107.07511, 2021.
\bibitem{ref3}\hypertarget{bib.ref3}{} Sadinle, Lei, Wasserman. \emph{Least Ambiguous Set-Valued Classifiers with Bounded Error Levels} (LAC/THR). JASA 2019. arXiv:1609.00451.
\bibitem{ref4}\hypertarget{bib.ref4}{} Romano, Sesia, Candès. \emph{Classification with Valid and Adaptive Coverage} (APS). NeurIPS 2020.
\bibitem{ref5}\hypertarget{bib.ref5}{} Angelopoulos, Bates, Malik, Jordan. \emph{Uncertainty Sets for Image Classifiers using Conformal Prediction} (RAPS). ICLR 2021. arXiv:2009.14193.
\bibitem{ref6}\hypertarget{bib.ref6}{} Ding, Angelopoulos, Bates, Jordan, Tibshirani. \emph{Class-Conditional Conformal Prediction with Many Classes} (clustered conformal). NeurIPS 2023. arXiv:2306.09335.
\bibitem{ref7}\hypertarget{bib.ref7}{} Vovk. \emph{Conditional Validity of Inductive Conformal Predictors} (Mondrian). ACML 2012.
\bibitem{ref8}\hypertarget{bib.ref8}{} Tibshirani, Barber, Candès, Ramdas. \emph{Conformal Prediction Under Covariate Shift.} NeurIPS 2019. arXiv:1904.06019.
\bibitem{ref9}\hypertarget{bib.ref9}{} Gibbs, Candès. \emph{Adaptive Conformal Inference Under Distribution Shift} (ACI). NeurIPS 2021. arXiv:2106.00170.
\bibitem{ref10}\hypertarget{bib.ref10}{} Kasa, Zhang, Yang, Taylor. \emph{Adapting Prediction Sets to Distribution Shifts Without Labels} (ECP/EACP). arXiv:2406.01416, 2024.
\bibitem{ref11}\hypertarget{bib.ref11}{} Silva-Rodríguez, Ben Ayed, Dolz. \emph{Conformal Prediction for Zero-Shot Models} (Conf-OT). CVPR 2025. arXiv:2505.24693.
\bibitem{ref12}\hypertarget{bib.ref12}{} Fillioux, Silva-Rodríguez, Ben Ayed, et al. \emph{Are foundation models for computer vision good conformal predictors?} arXiv:2412.06082, 2024 (v3 2026).
\bibitem{ref13}\hypertarget{bib.ref13}{} Alijani, Najjaran. \emph{WQLCP: Weighted Adaptive Conformal Prediction for Robust Uncertainty Quantification Under Distribution Shifts.} CVPR Workshops 2025. arXiv:2505.19587.
\bibitem{ref14}\hypertarget{bib.ref14}{} Radford et al. \emph{Learning Transferable Visual Models From Natural Language Supervision} (CLIP). ICML 2021. arXiv:2103.00020.
\bibitem{ref15}\hypertarget{bib.ref15}{} Cherti et al. \emph{Reproducible Scaling Laws for Contrastive Language-Image Learning} (OpenCLIP). CVPR 2023. arXiv:2212.07143.
\bibitem{ref16}\hypertarget{bib.ref16}{} Zhai et al. \emph{Sigmoid Loss for Language Image Pre-Training} (SigLIP). ICCV 2023. arXiv:2303.15343.
\bibitem{ref17}\hypertarget{bib.ref17}{} Recht et al. \emph{Do ImageNet Classifiers Generalize to ImageNet?} (ImageNet-V2). ICML 2019. arXiv:1902.10811.
\bibitem{ref18}\hypertarget{bib.ref18}{} Hendrycks et al. \emph{The Many Faces of Robustness: A Critical Analysis of Out-of-Distribution Generalization} (ImageNet-R). ICCV 2021. arXiv:2006.16241.
\bibitem{ref19}\hypertarget{bib.ref19}{} Hendrycks et al. \emph{Natural Adversarial Examples} (ImageNet-A). CVPR 2021. arXiv:1907.07174.
\bibitem{ref20}\hypertarget{bib.ref20}{} Wang et al. \emph{Learning Robust Global Representations by Penalizing Local Predictive Power} (ImageNet-Sketch). NeurIPS 2019. arXiv:1905.13549.
\bibitem{ref21}\hypertarget{bib.ref21}{} Hendrycks, Dietterich. \emph{Benchmarking Neural Network Robustness to Common Corruptions and Perturbations} (ImageNet-C). ICLR 2019. arXiv:1903.12261.
\bibitem{ref22}\hypertarget{bib.ref22}{} Geifman, El-Yaniv. \emph{Selective Classification for Deep Neural Networks.} arXiv:1705.08500, 2017.
\bibitem{ref23}\hypertarget{bib.ref23}{} Shi, Ghosh, Belkhouja, Doppa, Yan. \emph{Conformal Prediction for Class-wise Coverage via Augmented Label Rank Calibration} (RC3P). NeurIPS 2024. arXiv:2406.06818.
\bibitem{ref24}\hypertarget{bib.ref24}{} Liu, Huang, Ong. \emph{Conformal Prediction Meets Long-tail Classification.} arXiv:2508.11345, 2025.
\bibitem{ref25}\hypertarget{bib.ref25}{} Ding, Fermanian, Salmon. \emph{Conformal Prediction for Long-Tailed Classification.} arXiv:2507.06867, 2025.
\bibitem{ref26}\hypertarget{bib.ref26}{} Bairaktari, Wu, Wu. \emph{Kandinsky Conformal Prediction: Beyond Class- and Covariate-Conditional Coverage.} ICML 2025. arXiv:2502.17264.
\bibitem{ref27}\hypertarget{bib.ref27}{} Xie, Zhou, Liang, Favaro, Sesia. \emph{Conformal Inference for Open-Set and Imbalanced Classification.} arXiv:2510.13037, 2025.
\bibitem{ref28}\hypertarget{bib.ref28}{} Bozorgtabar, Mahapatra, Roy, Naseer, Razzak, Ge. \emph{LATA: Laplacian-Assisted Transductive Adaptation for Conformal Uncertainty in Medical VLMs.} arXiv:2602.17535, 2026.
\bibitem{ref29}\hypertarget{bib.ref29}{} Thopalli, Narayanaswamy, Thiagarajan. \emph{The Surprising Utility of Group Partitioning in Improving Conformal Prediction of Visual Classifiers under Distributional Shifts.} CVPR Workshops 2025.
\bibitem{ref30}\hypertarget{bib.ref30}{} Kasa, Taylor. \emph{Empirically Validating Conformal Prediction on Modern Vision Architectures Under Distribution Shift and Long-tailed Data.} arXiv:2307.01088, 2023.
\bibitem{ref31}\hypertarget{bib.ref31}{} Zhou, Fathony, Nguyen, Sesia. \emph{Audited Conformal Prediction for Classification under Unknown Distribution Shift.} arXiv:2606.14909, 2026.
\bibitem{ref32}\hypertarget{bib.ref32}{} Silva-Rodríguez, Ben Ayed, Dolz. \emph{Trustworthy Few-Shot Transfer of Medical VLMs through Split Conformal Prediction} (SCA-T). MICCAI 2025. arXiv:2506.17503.
\bibitem{ref33}\hypertarget{bib.ref33}{} Marani, Silva-Rodríguez, Ben Ayed, Vakalopoulou, Christodoulidis, Dolz. \emph{Class Adaptive Conformal Training} (CaCT). arXiv:2601.09522, 2026.
\bibitem{ref34}\hypertarget{bib.ref34}{} Braun, Holzmüller, Jordan, Bach. \emph{Conditional Coverage Diagnostics for Conformal Prediction.} arXiv:2512.11779, 2025.
\bibitem{ref35}\hypertarget{bib.ref35}{} Xu et al. \emph{Demystifying CLIP Data} (MetaCLIP). ICLR 2024. arXiv:2309.16671.
\bibitem{ref36}\hypertarget{bib.ref36}{} Gadre et al. \emph{DataComp: In search of the next generation of multimodal datasets.} NeurIPS 2023. arXiv:2304.14108.
\bibitem{ref37}\hypertarget{bib.ref37}{} Fang et al. \emph{Data Filtering Networks} (DFN). ICLR 2024. arXiv:2309.17425.
\bibitem{ref38}\hypertarget{bib.ref38}{} Barber, Candès, Ramdas, Tibshirani. \emph{The Limits of Distribution-Free Conditional Predictive Inference.} Information and Inference, 2021. arXiv:1903.04684.
\bibitem{ref39}\hypertarget{bib.ref39}{} Russakovsky et al. \emph{ImageNet Large Scale Visual Recognition Challenge.} IJCV 2015. arXiv:1409.0575.
\bibitem{ref40}\hypertarget{bib.ref40}{} Krause, Stark, Deng, Fei-Fei. \emph{3D Object Representations for Fine-Grained Categorization} (Stanford Cars). ICCV Workshops 2013.
\bibitem{ref41}\hypertarget{bib.ref41}{} Bossard, Guillaumin, Van Gool. \emph{Food-101: Mining Discriminative Components with Random Forests.} ECCV 2014.
\bibitem{ref42}\hypertarget{bib.ref42}{} Schuhmann, Beaumont, Vencu, Gordon, et al. \emph{LAION-5B: An Open Large-Scale Dataset for Training Next Generation Image-Text Models.} NeurIPS 2022 Datasets and Benchmarks. arXiv:2210.08402.
\bibitem{ref43}\hypertarget{bib.ref43}{} Chen, Wang, Changpinyo, et al. \emph{PaLI: A Jointly-Scaled Multilingual Language-Image Model.} ICLR 2023. arXiv:2209.06794.
\bibitem{ref44}\hypertarget{bib.ref44}{} Naganuma, Hataya, Yoshida, Mitliagkas. \emph{An Empirical Study of Pre-trained Model Selection for Out-of-Distribution Generalization and Calibration.} TMLR 2024. arXiv:2307.08187.
\end{thebibliography}
\end{document}